\documentclass[a4paper,fleqn]{cas-dc}
\usepackage{retailor}

\begin{document}

\let\WriteBookmarks\relax
\shorttitle{ReTailor}
\shortauthors{J.-Z. Liang et al.}

\title[mode=title]{ReTailor: A Residual Feature Tailoring Framework for Real-Time Cross-Sensor Pansharpening Adaptation }

\author[1]{Jing-Ze Liang}[orcid=0009-0004-1925-9466]
\fnmark[1]
\ead{liangjingze7@gmail.com}
\credit{Conceptualization, Methodology, Software, Validation,
Investigation, Writing -- original draft, Writing -- review \& editing}

\author[1]{Wei-Jie Xiao}[orcid=0009-0006-5125-7043]
\fnmark[1]
\ead{WeiJie.Xiao06@gmail.com}
\credit{Methodology, Validation, Investigation, Writing -- review \& editing}

\author[1]{Zi-Han Cao}[orcid=0000-0002-7532-2122]
\fnmark[1]
\ead{iamzihan666@gmail.com}
\credit{Validation, Investigation, Visualization, Writing -- review \& editing}

\author[1]{Yang-Dong Li}[orcid=0009-0003-2754-9319]
\ead{2025080901010@std.uestc.edu.cn}
\credit{Investigation, Data curation, Writing -- review \& editing}

\author[1]{Tian-Yi Hu}[orcid=0009-0006-2766-7456]
\ead{2025081001005@std.uestc.edu.cn}
\credit{Investigation, Data curation, Writing -- review \& editing}

\author[1,2]{Liang-Jian Deng}[orcid=0000-0003-3178-9772]
\cormark[1]
\ead{liangjian.deng@uestc.edu.cn}
\credit{Supervision, Project administration,
Writing -- review \& editing}

\affiliation[1]{
  organization={University of Electronic Science and Technology of China},
  city={Chengdu},
  postcode={611731},
  state={Sichuan},
  country={China}
}

\affiliation[2]{
  organization={Multi-Hazard Early Warning Key Laboratory of Sichuan Province},
  city={Chengdu},
  postcode={611731},
  state={Sichuan},
  country={China}
}

\cortext[cor1]{Corresponding author}
\fntext[fn1]{These authors contributed equally to this work.}

\begin{abstract}
Recent pansharpening models achieve strong fusion performance on trained sensors, yet often generalize poorly to unseen sensors. Existing cross-sensor adaptation is limited by target-sensor retraining, which requires target-sensor training pairs and a dedicated training stage, and by per-image zero-shot adaptation, which is computationally expensive due to whole-network optimization for each test image. To address these problems, we propose ReTailor, a real-time sequential cross-sensor adaptation framework. Only a lightweight residual Feature Tailor is rapidly optimized under self-supervised, physics-aware constraints to correct cross-sensor shifts without modifying the well-learned fusion prior. For efficient and effective deployment, ReTailor combines content-aware patch selection, which identifies a few informative patches per image for optimization at negligible cost to improve both quality and efficiency, with rolling residual bootstrap, which inherits sensor-related knowledge across the image stream to further improve quality. Extensive experiments on cross-sensor adaptation settings show that ReTailor mitigates cross-sensor degradation both effectively and efficiently. Moreover, we compare ReTailor with Low-Rank Adaptation (LoRA), a widely used method for fine-tuning, to demonstrate the advantages of our approach. We further evaluate the core principle of ReTailor on two additional remote sensing tasks: super-resolution pansharpening and image segmentation, where it consistently improves performance. In general, ReTailor provides a highly effective and efficient strategy for deploying pansharpening models under real-world cross-sensor degradation, while also offering useful insights into managing cross-sensor degradation in other remote sensing tasks. Code will be released at \url{https://github.com/JingZe-Liang/ReTailor}.
\end{abstract}


\begin{keywords}
 Cross-sensor pansharpening \sep
 Instantaneous adaptation \sep
 Continual test-time adaptation \sep
 Adaptive selection
\end{keywords}

\maketitle

\section{Introduction}
\label{sec:introduction}

\begin{figure}[pos=t]
    \vspace*{6pt}
    \centering
    \includegraphics[width=\columnwidth]{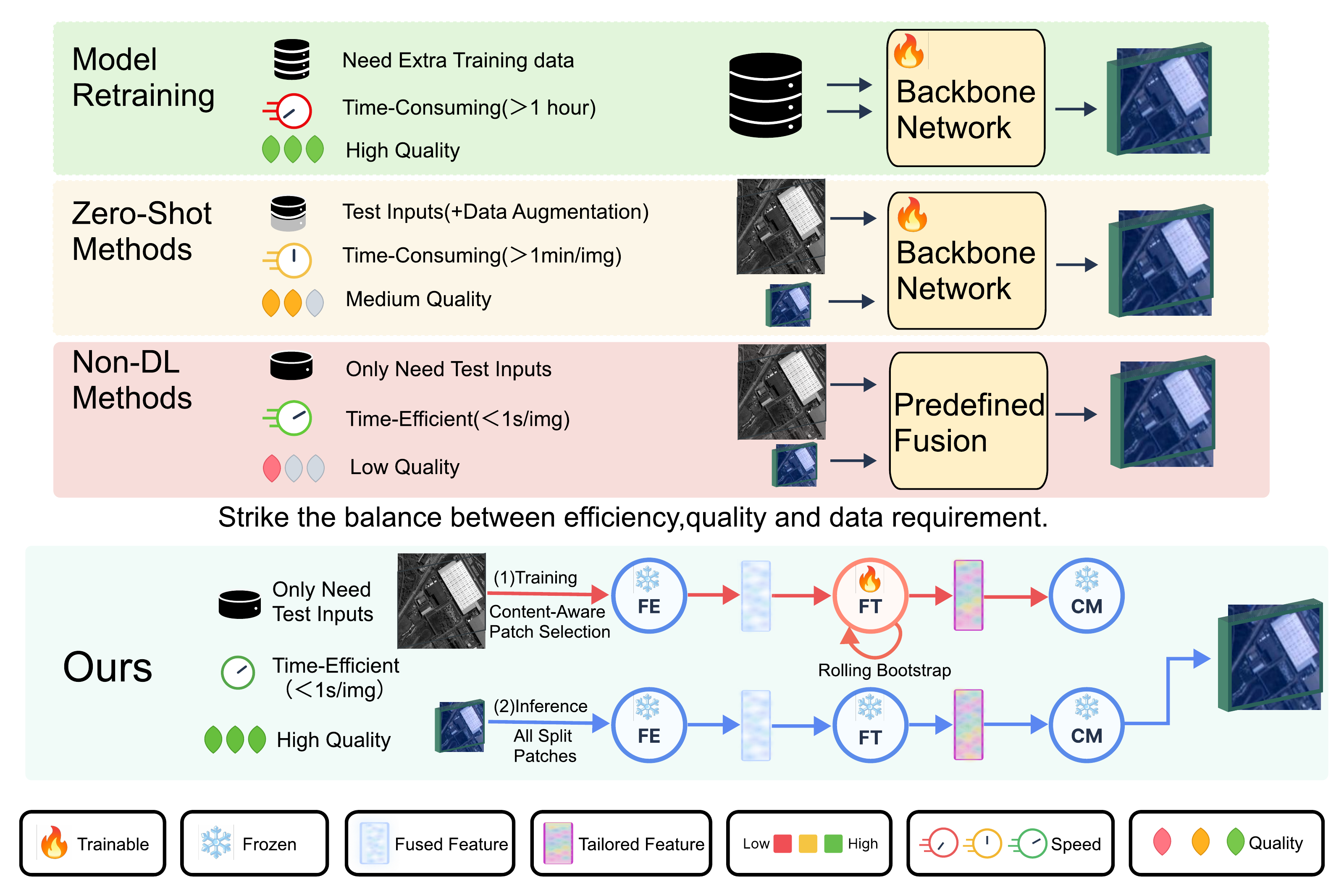}
    \vspace{-6pt}
    \caption{\textbf{Comparison of cross-sensor adaptation paradigms.} Existing methods (top) trade fusion quality against efficiency and ground-truth (GT) requirements, whereas ReTailor (bottom) achieves high-quality, real-time adaptation without GT while preserving the pretrained fusion prior.}
    \label{fig:cross-sensor-comparison}
\end{figure}
Pansharpening reconstructs a high-resolution multispectral (HRMS) image from the fine spatial details of a high-resolution panchromatic (PAN) image and the spectral information of a low-resolution multispectral (LRMS) image. Due to its importance in remote-sensing interpretation and downstream analysis, pansharpening has been well studied~\cite{deng2022benchmark,dadrassjavan2021review,duran2017survey}. Traditional methods mainly include component substitution, multiresolution analysis, and variational optimization~\cite{amolins2007wavelet}. Component-substitution and multiresolution-analysis methods transfer PAN details through explicit injection rules~\cite{tu2001ihs, tu2004fastihs,aiazzi2006mtf,cheng2015wavelet}, whereas variational methods formulate fusion as a regularized reconstruction problem~\cite{ballester2006pxs,fang2013variational,palsson2014tv,gogineni2018sparsity}. Deep-learning methods instead learn nonlinear mappings from PAN and LRMS observations to HRMS images. Representative architectures include convolutional neural networks~\cite{masi2016pnn,scarpa2018targetadaptive,deng2021fusionnet,xu2021gppnn}, Transformers~\cite{zhou2022panformer}, diffusion models~\cite{meng2023pandiff}, and state-space models~\cite{he2025panmamba}.
\begin{figure*}[t]
    \centering
    \includegraphics[width=\textwidth]{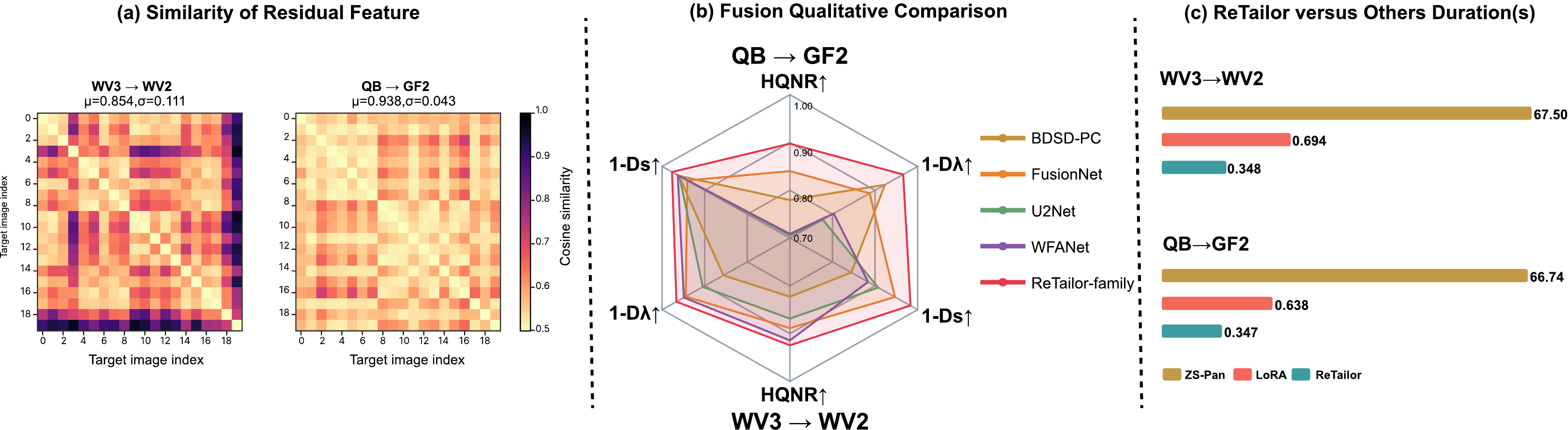}
    \caption{\textbf{Empirical overview of ReTailor.} (a) Pairwise cosine similarity between residual embeddings of independently adapted FusionNet Feature Tailors on 20 target images per transfer (mean $\pm$ std over 190 pairs). (b) Fusion quality (HQNR, $1-D_{\lambda}$, $1-D_s$; higher is better) on QB$\rightarrow$GF2 (upper axes) and WV3$\rightarrow$WV2 (lower axes); the ReTailor-family curve takes the best mean of FT-only and ReTailor over both backbones (the HQNR and $1-D_{\lambda}$ values are those of ReTailor$_{\mathrm{FusionNet}}$ alone, the best method in Table~\ref{tab:main-results}). (c) End-to-end per-image duration of ZS-Pan~\cite{cao2024zspan}, LoRA~\cite{hu2022lora}, and ReTailor (FusionNet); smaller is faster.}
    \label{3in1}

\end{figure*}
Despite this progress, the performance of existing pansharpening methods often degrades heavily in cross-sensor applications. Sensor-dependent discrepancies in spatial and spectral responses, noise patterns, and radiometric statistics can cause the learned fusion priors to break down when transferred across sensors~\cite{scarpa2018targetadaptive}. Existing solutions mainly follow two routes: target-sensor retraining and zero-shot adaptation. Retraining recalibrates a model to the target sensor distribution, but it requires target-sensor training pairs, either ground truth (GT) data or reduced-resolution pairs simulated from target observations, together with lengthy optimization. GT data are also difficult to acquire~\cite{scarpa2018targetadaptive}. Zero-shot methods avoid additional training by optimizing the model at inference time. However, most approaches still require tens of seconds to several minutes of processing for each input pair~\cite{rui2024variationalzs,cao2024zspan}. Traditional methods avoid both retraining and test-time optimization, but their limited nonlinear representation capacity generally results in lower fusion quality~\cite{amro2011survey,deng2022benchmark}. A systematic comparison of these solution paradigms with our method is presented in Fig.~\ref{fig:cross-sensor-comparison}. In summary, an effective adaptation method should be \textit{fast, require little or no GT data, and maintain high fusion quality.}

\begin{figure*}[t]
    \centering
    \includegraphics[width=\textwidth]{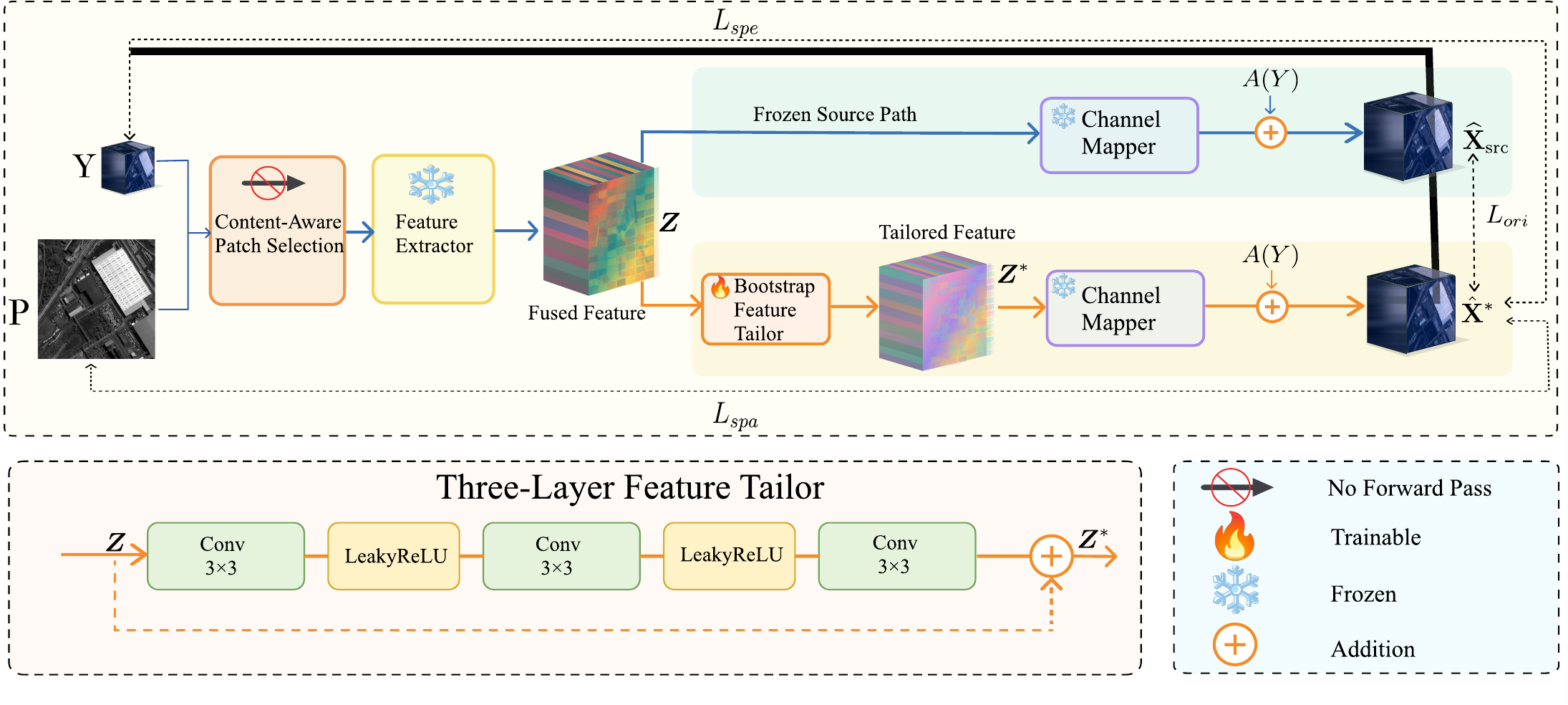}
    \caption{\textbf{Deployment pipeline of ReTailor.} Content-aware selection picks informative PAN/LRMS patches. The frozen Feature Extractor produces latent features $\mathbf{Z}$, the trainable Feature Tailor adds a residual correction to give $\mathbf{Z}^{*}$, and the frozen Channel Mapper outputs the source and tailored predictions. Together with the observations and target-sensor modulation transfer function (MTF) priors, these form the spectral, spatial, and frozen-source losses that update only the Tailor. The lower panel shows the three-layer Tailor.}
    \label{fig:retailor-framework}
\end{figure*}

To meet these requirements, we propose ReTailor, a cross-sensor adaptation framework as illustrated in Fig.~\ref{fig:retailor-framework}. ReTailor effectively mitigates cross-sensor pansharpening degradation, outperforming existing baselines in fusion quality (Fig.~\ref{3in1}(b)) while reducing per-image adaptation latency to a fraction of a second (Fig.~\ref{3in1}(c)). Given a stream of target PAN/LRMS pairs, each pair is first partitioned into spatially aligned patches. Adaptation of the seed image (first image) uses all patches, whereas content-aware patch selection determines the adaptation subset for each subsequent image with negligible computational overhead. The corresponding patch set is then used to adapt the pretrained fusion model. Between the frozen Feature Extractor and Channel Mapper, both defined in Sec.~\ref{sec:residual-feature-tailoring}, we insert a lightweight residual Feature Tailor. Using the selected informative patch set, the Feature Tailor is updated to correct the cross-sensor degradation under self-supervised, physics-aware constraints, while the Feature Extractor and Channel Mapper remain frozen. After adaptation, all candidate patches are encoded by the frozen Feature Extractor, corrected by the updated Feature Tailor, and mapped by the frozen Channel Mapper into HRMS patches, which are then stitched into the full-resolution output.

To identify fewer yet more informative patches as quickly as possible, while retaining sufficient information for rapid and generalizable adaptation to cross-sensor degradation rather than biased overfitting, ReTailor leverages two complementary and mutually constraining cues. Within-patch PAN variation provides spatial-detail cues, whereas inter-band LRMS variation provides complementary spectral cues; both are available directly from the PAN/LRMS pair~\cite{yilmaz2022survey}. As target images are processed sequentially, rolling residual bootstrap propagates the learned Tailor correction across images rather than restarting adaptation from the default initialization for each image. Supported by the ablation and the analysis in Sec.~\ref{sec:residual-probe}, this cross-image knowledge propagation preserves previously learned sensor-related corrections and improves per-image adaptation quality. We further verify that it remains stable over long image sequences (also in Sec.~\ref{sec:long-sequence-stability}). Fig.~\ref{3in1}(a) provides the visual evidence for this design.

Following this workflow, these components work together with distinct roles to enable efficient and effective cross-sensor adaptation. Content-aware patch selection prioritizes patches that are most informative for correcting cross-sensor degradation while avoiding uninformative samples that may lead to biased adaptation, with negligible additional computational overhead. The residual Feature Tailor then extracts corrective information from the selected patches into lightweight residual adjustments, enabling rapid cross-sensor correction. Rolling residual bootstrap further carries forward effective corrections learned from preceding images, exploiting their shared cross-sensor degradation characteristics to improve subsequent adaptation. \textit{Together, ReTailor delivers strong cross-sensor fusion performance, achieving up to 0.9245 HQNR while keeping end-to-end latency below 0.57 s (selection + adaptation + pansharpening) per image across all evaluated settings.}

We evaluate ReTailor against representative traditional and learning-based pansharpening methods on QB$\rightarrow$GF2 and WV3$\rightarrow$WV2. ReTailor$_{\mathrm{FusionNet}}$ achieves the highest HQNR and the lowest spectral distortion $D_{\lambda}$ on both transfers, reaching HQNR values of 0.8976 and 0.9245, respectively. These gains are achieved with only about six selected patches and 0.12--0.20 s of adaptation per image, demonstrating that high-quality cross-sensor correction can be obtained with a very limited adaptation budget. This efficiency is further confirmed by fair comparisons with LoRA, where ReTailor reduces mean end-to-end latency by almost 50\% while consistently lowering $D_{\lambda}$ across all evaluated settings. Detailed ablation studies validate the contribution of each proposed component. Beyond standard pansharpening, we further examine the ReTailor principle on two additional cross-sensor remote-sensing tasks, namely arbitrary-scale super-resolution pansharpening and semantic segmentation, where it consistently remains effective.

The main contributions of this work are summarized as follows:
\begin{enumerate}
    \item We develop \textbf{ReTailor}, an efficient cross-sensor adaptation framework that preserves the pretrained fusion prior, rapidly corrects cross-sensor degradation, and maintains high fusion quality. 
    
    \item We carefully design a fully tensorized and highly parallel content-aware patch selection method, a lightweight residual Feature Tailor, and a rolling residual bootstrap strategy, which play complementary roles and jointly benefit the ReTailor framework.

    \item We conduct extensive experiments on cross-sensor pansharpening datasets covering both 4-band and 8-band multispectral imagery, with comparisons against representative traditional and learning-based methods as well as a parameter-matched LoRA baseline. We further extend the core principle of ReTailor to arbitrary-scale super-resolution pansharpening and cross-sensor semantic segmentation, highlighting its broader potential for cross-sensor adaptation in remote sensing.
\end{enumerate}

\section{Related Work}
\label{sec:related_work}

\subsection{Learning-Based Pansharpening}

Deep learning methods have achieved remarkable performance in pansharpening by learning nonlinear mappings from PAN and LRMS observations to HRMS images~\cite{xing2018metric,jiang2020differential,zhang2021gtp,he2023selfsupervised}. Early approaches were mainly based on convolutional neural networks (CNNs). PNN pioneered direct end-to-end learning for pansharpening from PAN and upsampled LRMS inputs~\cite{masi2016pnn}, while PanNet incorporated high-pass domain priors to explicitly enhance spatial-detail injection~\cite{yang2017pannet}. Subsequent CNN-based methods further improved feature interaction and detail reconstruction, as exemplified by FusionNet~\cite{deng2021fusionnet,cui2025predictive} and GPPNN~\cite{xu2021gppnn,han2024tempansharpening}.

More recent studies have explored architectures beyond conventional CNNs. PanFormer employs Transformer-based modeling to capture long-range spatial--spectral dependencies~\cite{zhou2022panformer}. PanDiff formulates pansharpening as a diffusion-based reconstruction process~\cite{meng2023pandiff}, while Pan-Mamba introduces state-space modeling for efficient long-range and cross-modal interaction~\cite{he2025panmamba}. Despite these architectural advances, most learning-based methods are trained under sensor-specific data distributions, and their performance often degrades when the spectral responses, spatial characteristics, or imaging statistics of the deployment sensor differ from those encountered during training.

\subsection{Cross-Sensor Pansharpening Strategies}

Applying a pansharpening model trained on one sensor to another remains challenging because sensor-dependent spatial-spectral responses, noise characteristics, and radiometric distributions may violate the learned fusion prior, resulting in substantial cross-sensor degradation~\cite{scarpa2018targetadaptive,volpi2015spectral,luo2022domain,qiu2023crosssr,xin2026erft}. Existing solutions can be broadly grouped into three categories:

\textbf{(1) Model Retraining.}
Retraining and fine-tuning methods update all or part of a pretrained pansharpening model using data acquired from the target sensor~\cite{shen2023continual}. Target-adaptive CNN-based pansharpening, for example, constructs reduced-resolution training pairs from target observations and recalibrates the learned fusion mapping according to the target acquisition characteristics~\cite{scarpa2018targetadaptive}. More recent sensor-transfer approaches reduce this cost by selecting a compact subset of target-domain samples and updating only parameters that are sensitive to the sensor shift~\cite{xia2026swift}. Although effective, these methods require target-sensor training pairs, either GT samples or reduced-resolution pairs simulated under Wald's protocol~\cite{wald1997fusion}, and an additional, often computationally demanding and complex training stage, with the pretrained model being partially or fully updated.

\textbf{(2) Zero-Shot Methods.}
Zero-shot methods optimize a pretrained pansharpening model during inference without requiring target GT supervision~\cite{cao2024zspan,rui2024variationalzs}. In remote sensing, these methods typically construct self-supervised objectives from the observed LRMS--PAN pair and exploit the sensor observation process to adapt the model to the current input. They avoid the need for additional target-sensor training data while maintaining relatively high fusion quality. Parameter-efficient techniques can also be incorporated into this zero-shot setting. For example, LoRA freezes the pretrained model weights and represents the update using trainable low-rank factors~\cite{hu2022lora}, thereby restricting inference-time optimization to a small number of parameters. When optimized separately for each target image under a self-supervised objective, LoRA serves as a parameter-efficient zero-shot adaptation mechanism. Despite their effectiveness, conventional zero-shot methods often require tens of seconds to several minutes to process a single LRMS--PAN input.

\textbf{(3) Traditional Methods.}
Traditional methods based on component substitution, multiresolution analysis, or variational optimization require neither target-sensor training nor inference-time network adaptation. Component substitution and multiresolution analysis methods inject PAN details through predetermined transformations, whereas variational methods reconstruct the fused image under explicit observation models and regularization constraints~\cite{tu2001ihs,tu2004fastihs,aiazzi2007gsa,otazu2005wisper}. These methods are generally computationally efficient and can operate directly on a single LRMS--PAN pair. However, their predefined fusion rules and limited nonlinear representation capacity restrict their fusion performance.

In summary, existing strategies face a trade-off among fusion quality, training-data requirements, and computational efficiency. Model retraining can achieve strong target-sensor performance but requires target-sensor training pairs and a dedicated training stage, whereas zero-shot methods remove the need for such training pairs at the cost of substantial per-image inference-time computation. Traditional methods are efficient and training-free but their performance is limited by predefined fusion mechanisms. By contrast, ReTailor addresses this trade-off by providing an efficient, effective, and practical solution to adapt pretrained pansharpening models to unseen sensors.

\section{Method}
\label{sec:method}
This section presents ReTailor. Section~\ref{sec:framework-overview} gives the overall pipeline and notation. Section~\ref{sec:residual-feature-tailoring} describes the residual Feature Tailor inserted into a frozen backbone, and Section~\ref{sec:adaptation-objective} defines the physics-aware objective that optimizes it without target HRMS. Section~\ref{sec:content-aware-selection} introduces content-aware patch selection, which builds a compact and high-value adaptation set (complexity in Appendix~\ref{app:selection-complexity}), and Section~\ref{sec:rolling-bootstrap} introduces rolling residual bootstrap, which lets each image inherit the sensor-related correction learned by its predecessors.

\subsection{Framework Overview}
\label{sec:framework-overview}

As shown in Fig.~\ref{fig:retailor-framework}, ReTailor combines three components: a residual Feature Tailor that corrects target-sensor representation shifts, content-aware patch selection that concentrates adaptation on informative observations, and rolling residual bootstrap that passes the previously learned sensor-related correction along the target-image stream.

For the $t$-th target image, the aligned LRMS and PAN observations are partitioned into $N_c$ spatially corresponding candidate patch pairs. We collect these candidates in the ordered collection
\begin{equation*}
    \mathcal{U}^{t}
    =
    \left[
        (\mathbf{Y}_{i}^{t},\mathbf{P}_{i}^{t})
    \right]_{i=1}^{N_c},
\end{equation*}
where $t$ indexes the image-processing order and $i$ indexes candidate patches within image $t$. Adaptation and inference use different observations. Let $\mathcal{S}^{t}\subseteq\{1,\ldots,N_c\}$ denote the adaptation index set, $\Omega^{t}=\mathcal{U}^{t}[\mathcal{S}^{t}]$ the corresponding selected observations, and $M^{t}=|\mathcal{S}^{t}|$ the number of selected patches. The seed image uses all candidates to establish an initial target-sensor correction, whereas each subsequent image uses content-aware patch selection to construct a compact adaptation set. Selection thus shrinks only the adaptation set; inference still covers the full image.

ReTailor decomposes a compatible pretrained pansharpening backbone into a frozen Feature Extractor $F_{\mathrm{ext}}$ and a frozen Channel Mapper $F_{\mathrm{map}}$, and inserts a residual Feature Tailor $G_{\phi}$ at their fused-feature interface. Only the Tailor parameters $\phi$ are optimized on $\Omega^{t}$ under the physics-aware adaptation objective. For the seed image ($t=0$), adaptation starts from the zero-output initialization $\phi_{\mathrm{init}}$ and uses all $N_c$ candidates as a one-time bootstrap. For each later image $t\geq1$, adaptation starts from the preceding state $\phi^{t-1}$ and refines it using the selected observations, producing the current Tailor state $\phi^{t}$.

After adaptation, $\phi^{t}$ is fixed and applied to all $N_c$ candidate patches of the current image. The frozen Feature Extractor first produces their fused representations, the Feature Tailor applies the learned residual correction, and the frozen Channel Mapper reconstructs the corresponding HRMS patches. These predictions are finally restored to their spatial order and stitched into the full-resolution output $\widehat{\mathbf{X}}^{t}$.

\subsection{Residual Feature Tailoring}
\label{sec:residual-feature-tailoring}

Residual Feature Tailoring preserves the source fusion pathway while adapting only its fused representation. Suppressing image and patch indices, the frozen Feature Extractor produces the fused latent tensor $\mathbf{Z}=F_{\mathrm{ext}}(\mathbf{Y},\mathbf{P};\theta_{\mathrm{ext}})$, and the frozen Channel Mapper produces the source prediction $\widehat{\mathbf{X}}_{\mathrm{src}}=F_{\mathrm{map}}(\mathbf{Z};\theta_{\mathrm{map}})+\mathcal{A}(\mathbf{Y})$. Here, $\theta_{\mathrm{ext}}$ and $\theta_{\mathrm{map}}$ are the frozen parameters of the Feature Extractor and Channel Mapper, respectively, and $\mathcal{A}(\mathbf{Y})$ is the fixed, non-trainable LRMS shortcut added to the Channel Mapper output to align it with the HRMS prediction space; it is $\mathbf{0}$ for a backbone without such a shortcut.

The Feature Tailor predicts an additive correction of this fused tensor; its implementation is detailed in the lower panel of Fig.~\ref{fig:retailor-framework}:
\begin{align}
    \mathbf{Z}^{*}
    &= \mathbf{Z}+G_{\phi}(\mathbf{Z}),
    \notag\\
    \widehat{\mathbf{X}}_{*}
    &= F_{\mathrm{map}}
    \left(\mathbf{Z}^{*}(\phi);\theta_{\mathrm{map}}\right)
    + \mathcal{A}(\mathbf{Y}),
    \label{eq:tailored-output}
\end{align}
where $G_{\phi}$ is the trainable Feature Tailor, $\mathbf{Z}^{*}$ is the corrected fused tensor, and $\widehat{\mathbf{X}}_{*}$ is the tailored prediction. Task-specific insertion interfaces and channel widths are summarized in Table~\ref{tab:retailor-configurations}. The modular decomposition underlying this insertion choice and the corresponding placement experiment are detailed in Appendix~\ref{app:ft-placement}.

\subsection{Physics-Aware Self-Supervised Adaptation Objective}
\label{sec:adaptation-objective}

At full-resolution deployment, target HRMS supervision is unavailable. ReTailor therefore adapts the Feature Tailor using only the selected observations $\Omega^{t}$, their frozen-source predictions, and the target-sensor MTF kernels $\mathbf{B}_{\mathrm{MS}}$ and $\mathbf{B}_{\mathrm{P}}$~\cite{aiazzi2006mtf}. With the adaptation index set $\mathcal{S}^{t}$ defined in Section~\ref{sec:framework-overview}, the objective is
\begin{equation}
    \resizebox{0.90\columnwidth}{!}{$\displaystyle
        \mathcal{L}_{\mathrm{adapt}}^{t}\!\left(\phi;\mathcal{S}^{t}\right)
        =\frac{1}{|\mathcal{S}^{t}|}\sum_{i\in\mathcal{S}^{t}}
        \Bigl[\eta_{\mathrm{spe}}\mathcal{L}_{\mathrm{spe}}^{t,i}(\phi)
        +\eta_{\mathrm{spa}}\mathcal{L}_{\mathrm{spa}}^{t,i}(\phi)
        +\eta_{\mathrm{ori}}\mathcal{L}_{\mathrm{ori}}^{t,i}(\phi)\Bigr]
    $},
    \label{eq:retailor-adaptation-objective}
\end{equation}
where $\mathcal{L}_{\mathrm{spe}}^{t,i}$, $\mathcal{L}_{\mathrm{spa}}^{t,i}$, and $\mathcal{L}_{\mathrm{ori}}^{t,i}$ are the spectral, spatial, and frozen-source consistency losses on patch $i$ of image $t$, $\eta_{\mathrm{spe}}$, $\eta_{\mathrm{spa}}$, and $\eta_{\mathrm{ori}}$ balance their contributions, and only $\phi$ is updated. The image, patch, and parameter indices are omitted below for readability.

For each selected patch, two observation-consistency terms are combined with a frozen-source regularizer:
\begin{align*}
    \mathcal{L}_{\mathrm{spe}}
    &=
    \left\|
        \mathcal{D}_{\mathrm{MS}}^{(b)}
        \!\left(\widehat{\mathbf{X}}_{*};\mathbf{B}_{\mathrm{MS}}\right)
        -
        \mathbf{Y}
    \right\|_{1},
    \\
    \mathcal{L}_{\mathrm{spa}}
    &=
    \left\|
        \widehat{\mathbf{X}}_{*}
        -
        \left(\widehat{\mathbf{X}}_{*}*\mathbf{B}_{\mathrm{MS}}\right)
        \odot
        \left[
            \widehat{\mathbf{P}}
            \oslash
            \left(
                \widehat{\mathbf{P}}*\mathbf{B}_{\mathrm{P}}
            \right)
        \right]
    \right\|_{1},
    \\
    \mathcal{L}_{\mathrm{ori}}
    &=
    \left\|
        \widehat{\mathbf{X}}_{*}
        -
        \widehat{\mathbf{X}}_{\mathrm{src}}
    \right\|_{1}.
\end{align*}

\noindent\textbf{Spectral consistency.}
The spectral term requires the tailored prediction to reproduce the observed LRMS patch after target-sensor degradation, which follows the consistency property of Wald's protocol~\cite{wald1997fusion} and the spectral-preservation objectives of unsupervised pansharpening~\cite{ma2020pangan,cao2024zspan}. Here, $b$ indexes the compatible frozen backbone, and $\mathcal{D}_{\mathrm{MS}}^{(b)}(\cdot;\mathbf{B}_{\mathrm{MS}})$ applies the target-sensor MS MTF blur and backbone-compatible resampling to a predicted HRMS patch.

\noindent\textbf{Spatial consistency.}
The spatial term constrains the tailored prediction using the relative high-frequency modulation observed in the PAN patch, in the spirit of contrast-based (ratio) detail injection~\cite{aiazzi2006mtf,vivone2014contrast} and the spatial-preservation objectives of unsupervised pansharpening~\cite{ma2020pangan,cao2024zspan}. The broadcast PAN patch $\widehat{\mathbf{P}}$ is represented in the HRMS channel space, and $\mathbf{B}_{\mathrm{P}}$ is the target-sensor PAN MTF kernel.

\noindent\textbf{Frozen-source consistency.}
The frozen-source term regularizes the tailored prediction toward $\widehat{\mathbf{X}}_{\mathrm{src}}$, which is independent of $\phi$, discouraging the observation-driven update from drifting away from the pretrained fusion prior.

The notation $\|\cdot\|_{1}$ denotes the mean absolute value, while $*$, $\odot$, and $\oslash$ denote convolution, element-wise multiplication, and element-wise division, respectively.

Because adaptation runs for only a few gradient steps on $\mathcal{S}^{t}$, fusion quality hinges on whether the selected patches provide sufficiently informative and complementary spatial--spectral constraints---the question taken up by the content-aware patch selection in Section~\ref{sec:content-aware-selection}.

\subsection{Content-Aware Patch Selection}
\label{sec:content-aware-selection}

Adapting on a subset of patches saves time, but the subset must be chosen well: what matters is not only how many patches are kept but which ones constrain the Tailor update. A fixed random subset ignores patch content and may include patches that carry little information about the cross-sensor shift. Content-aware scores such as the Tailor-parameter gradient norm~\cite{paul2021datadiet} or the loss reduction after a trial update~\cite{killamsetty2021glister} would require at least one extra forward and backward pass per candidate, which defeats the purpose of fast adaptation.

ReTailor instead scores each candidate from PAN/LRMS cues that are available before adaptation and need no forward or gradient computation, so selection takes only a few milliseconds. To keep high-scoring but spatially clustered candidates from dominating the budget, regional discounting lowers the priority of candidates whose coarse region is already represented. The resulting set carries strong adaptation evidence with broad spatial coverage.

\begin{figure*}[t]
    \centering
    \includegraphics[width=\textwidth]{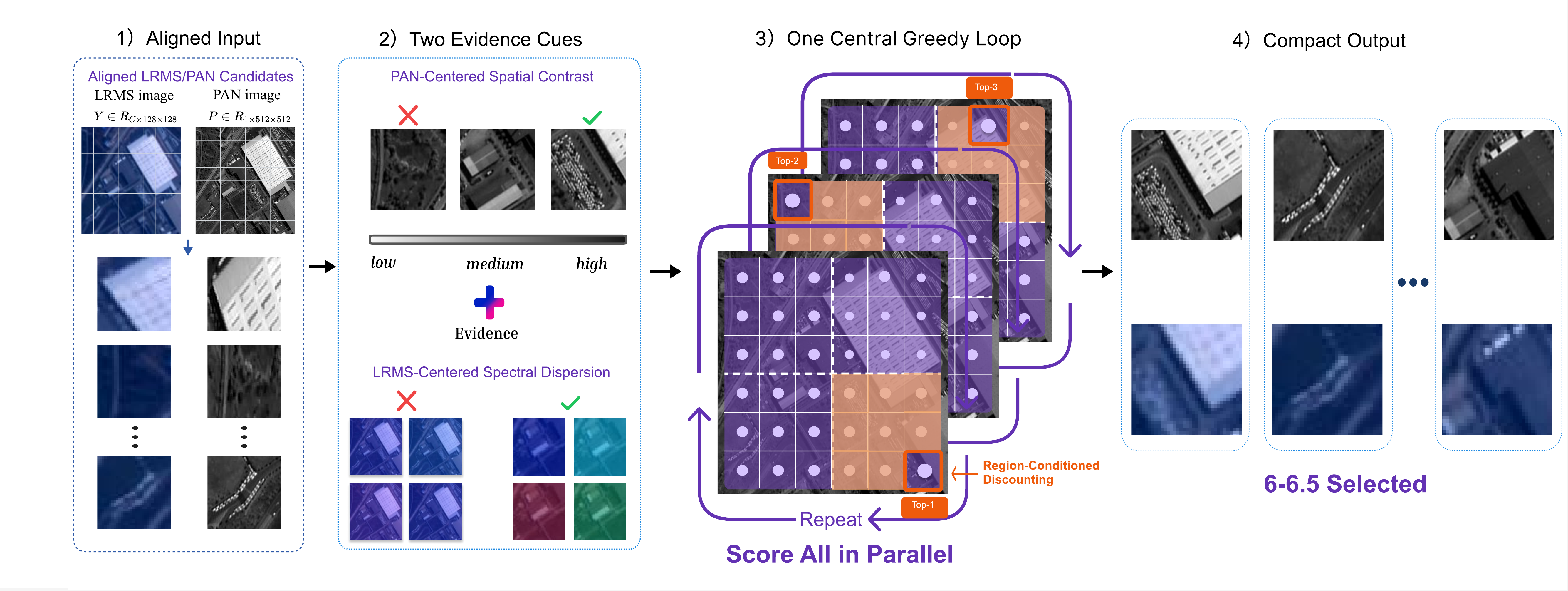}
    \caption{\textbf{Content-aware patch selection.} PAN spatial variation and LRMS inter-band variation of the $N_c$ candidates are combined into a PAN-anchored score, discounted by regional occupancy, and greedily admitted until relative stopping ($\tau$) yields $M^{t}$ patches for adaptation; inference still covers all $N_c$ candidates.}
    \label{fig:content-aware-selector}
\end{figure*}

As summarized in Fig.~\ref{fig:content-aware-selector}, for each later image $t\geq1$, the selector maps $\mathcal{U}^{t}$ to a nonempty set $\mathcal{S}^{t}$, with $M^{t}$ determined by $\tau$. The seed image ($t=0$) instead uses all $N_c$ candidates, as specified in Section~\ref{sec:rolling-bootstrap}.

\subsubsection{PAN/LRMS variation measures for adaptation patch selection}

For each aligned candidate patch in a target image, the selector computes two scalar variation measures that provide complementary adaptation evidence: $g_i$ quantifies within-patch PAN spatial variation, whereas $s_i$ quantifies inter-band LRMS spectral variation. Let $L=hw$ denote the number of PAN pixels, $L_s=h_sw_s$ the number of LRMS spatial samples per band, and $C_{\mathrm{MS}}$ the number of LRMS spectral bands. We represent the PAN patch as $\mathbf{p}_{i}=\operatorname{vec}(\mathbf{P}_{i})\in\mathbb{R}^{L}$ and the LRMS patch as $\mathbf{Y}_{i}\in\mathbb{R}^{C_{\mathrm{MS}}\times L_s}$. To remove a global PAN brightness offset and the LRMS component shared across spectral bands, ReTailor applies centering projections before computing these measures:
{\small
\begin{align*}
    \mathbf{H}_{L}
    &=
    \mathbf{I}_{L}
    -
    \frac{1}{L}
    \mathbf{1}_{L}
    \mathbf{1}_{L}^{\top},
    &
    g_i
    &=
    \frac{1}{\sqrt{L}}
    \left\|
        \mathbf{H}_{L}\mathbf{p}_{i}
    \right\|_{2},\\
    \mathbf{H}_{\mathrm{MS}}
    &=
    \mathbf{I}_{C_{\mathrm{MS}}}\!
    -
    \frac{1}{C_{\mathrm{MS}}}\!
    \mathbf{1}_{C_{\mathrm{MS}}}
    \mathbf{1}_{C_{\mathrm{MS}}}^{\top},
    &
    s_i
    &=
    \frac{1}{C_{\mathrm{MS}}\!L_s}
    \left\|
        \mathbf{H}_{\mathrm{MS}}
        \mathbf{Y}_{i}
    \right\|_{1,1},
\end{align*}
}
where $\mathbf{I}_{n}$ and $\mathbf{1}_{n}$ are the $n\times n$ identity matrix and length-$n$ all-ones vector, $\|\cdot\|_{2}$ is the Euclidean norm, and $\|\cdot\|_{1,1}$ is the sum of absolute matrix entries. The projection $\mathbf{H}_{L}$ centers PAN values over spatial positions, whereas $\mathbf{H}_{\mathrm{MS}}$ centers each LRMS spatial sample over spectral bands. Consequently, $g_i$ and $s_i$ are invariant to the corresponding additive common modes and can be computed in parallel for all candidates.

\subsubsection{PAN-anchored evidence composition and coarse-region mapping}

Because the PAN and LRMS cues $g_i$ and $s_i$ have different numerical ranges, ReTailor independently applies within-image min--max normalization to each cue. For $x\in\{g,s\}$, $\widetilde{x}_i$ denotes the normalized value of $x_i$, with the minimum and maximum computed over $j\in\{1,\ldots,N_c\}$.

The normalized PAN spatial-variation measure $\widetilde{g}_i$ is used as the score anchor (supported by ablation in Sec.~\ref{sec:evidence-composition}) because it quantifies within-patch variation at the target spatial scale, providing a direct proxy for the high-frequency detail constrained by the PAN observation. This is consistent with pansharpening practice, where detail injection is gated by local statistics of the PAN/MS pair~\cite{aiazzi2002context} or modulated by local PAN contrast~\cite{vivone2014contrast}, and learning-based methods operate on the high-pass PAN content~\cite{yang2017pannet}. The normalized LRMS inter-band spectral-variation measure $\widetilde{s}_i$ captures complementary spectral evidence from the lower-resolution LRMS observation. Spectral fidelity is commonly assessed through inter-band relationships~\cite{alparone2008qnr}, and band-dependent injection models treat each band individually~\cite{vivone2019robust}, so patches with strong inter-band variation constrain the spectral loss most. We therefore use a one-sided composition in which LRMS evidence supplements the PAN-based score only when $\widetilde{s}_i$ exceeds $\widetilde{g}_i$:

\begin{equation}
    b_i
    =
    \widetilde{g}_i
    +
    \lambda_c
    [\widetilde{s}_i-\widetilde{g}_i]_{+},
    \qquad 0\leq\lambda_c\leq1,
    \label{eq:base-utility}
\end{equation}
where $[x]_{+}=\max(x,0)$ and $\lambda_c$ controls the LRMS supplement. The positive-part term treats the LRMS cue as a bounded supplement rather than a competing score: it leaves the PAN baseline unchanged when $\widetilde{s}_i\leq\widetilde{g}_i$ and increases the score when the LRMS cue indicates additional spectral distinctiveness.

The base score in Eq.~\eqref{eq:base-utility} ranks candidates solely by PAN/LRMS evidence and does not account for whether high-scoring candidates are spatially clustered. ReTailor therefore divides the image into $N_r$ fixed coarse regions. Each candidate is assigned to the coarse region in which its patch center lies, and $r(i)\in\{1,\ldots,N_r\}$ denotes that region index. During selection, candidates in regions already represented in the selected set receive lower effective scores. The selector configuration is $\Xi_{\mathrm{sel}}=(\lambda_c,\lambda_r,N_r,\tau)$, where $\lambda_r$ controls regional discounting and $\tau\in[0,1]$ controls relative stopping. The settings used in ReTailor are specified in Appendix~\ref{app:implementation-details}.

\subsubsection{Tensorized region-aware selection and relative stopping}

To exploit informative patches for cross-sensor adaptation while avoiding regional overfitting, ReTailor combines parallelized region-aware greedy selection with $\tau$-controlled relative stopping. Candidates are first stably sorted by decreasing $b_i$:
\begin{align*}
    \boldsymbol{\pi}
    &=
    \operatorname{argsort}_{\downarrow}^{\mathrm{stable}}
    \!\left((b_i)_{i=1}^{N_c}\right),\\
    \bar{b}_{\ell}
    &=
    b_{\pi_{\ell}},
    \qquad
    \bar{r}_{\ell}
    =
    r(\pi_{\ell}),
    \quad
    \ell\in\{1,\ldots,N_c\},
\end{align*}
where $\boldsymbol{\pi}=(\pi_1,\dots,\pi_{N_c})^{\top}$ is the permutation
induced by the stable descending sort, whose $\ell$-th entry $\pi_\ell$
records the original index of the candidate ranked $\ell$-th, i.e.,
$\bar{b}_\ell=b_{\pi_\ell}$ and $\bar{r}_\ell=r_{\pi_\ell}$; equal scores
retain their original index order. The search starts from:
\begin{equation*}
    \mathcal{S}^{(0)}=\varnothing,
    \qquad
    \mathbf{a}^{(0)}=\mathbf{1}_{N_c},
    \qquad
    \mathbf{n}^{(0)}=\mathbf{0}_{N_r},
\end{equation*}
where the superscript denotes the admission round of the greedy search.
After $m$ admissions, $\mathcal{S}^{(m)}$ is the selected set,
$\mathbf{a}^{(m)}$ is the availability mask over sorted positions, and
$\mathbf{n}^{(m)}$ records the number of selected candidates in each
coarse region. The displayed initialization is the $m=0$ state, before
any candidate is admitted.

At round $m<N_c$, ReTailor gathers each candidate's current regional occupancy and computes its masked effective score. The sorted base scores and region indices are collected as $\bar{\mathbf{b}}=(\bar{b}_{1},\ldots,\bar{b}_{N_c})^{\top}$ and $\bar{\mathbf{r}}=(\bar{r}_{1},\ldots,\bar{r}_{N_c})^{\top}$, respectively:
\begin{align*}
    \mathbf{d}^{(m)}
    &=
    \mathbf{1}_{N_c}
    +
    \lambda_r
    \mathbf{n}^{(m)}[\bar{\mathbf{r}}],\\
    \mathbf{e}^{(m)}
    &=
    \operatorname{where}
    \!\left(
        \mathbf{a}^{(m)}=1,
        \bar{\mathbf{b}}\oslash\mathbf{d}^{(m)},
        -\infty
    \right),
    \qquad
    \lambda_r\geq0,
\end{align*}
where $\mathbf{d}^{(m)}$ is the region-discount vector and $\mathbf{e}^{(m)}$ is the masked effective-score vector used for admission. The indexed vector $\mathbf{n}^{(m)}[\bar{\mathbf{r}}]$ gathers the current regional count for each sorted candidate, so $\lambda_r\mathbf{n}^{(m)}[\bar{\mathbf{r}}]$ increases the discount in proportion to previous admissions from the same region. The operator $\operatorname{where}(c,x,y)$ acts element-wise, returning $x$ when $c$ is true and $y$ otherwise; hence, an available candidate receives $e_{\ell}^{(m)}=\bar{b}_{\ell}/d_{\ell}^{(m)}$, whereas an admitted candidate is assigned $-\infty$ and excluded from later rounds. Thus, $\mathbf{e}^{(m)}$ combines the base scores in $\bar{\mathbf{b}}$ with the regional discount in $\mathbf{d}^{(m)}$. Ranking the available candidates by $\mathbf{e}^{(m)}$ favors candidates with strong adaptation evidence and complementary regional coverage, thereby limiting repeated selection from the same region and reducing regional overfitting, as verified in Section~\ref{sec:regional-discount}.

At round $m$, the selector chooses the available position with the largest effective score, $\ell_{m+1}=\min\bigl(\arg\max_{\ell}e_{\ell}^{(m)}\bigr)$, and maps it to the original candidate index $i_{m+1}=\pi_{\ell_{m+1}}$. The first candidate is admitted unconditionally and sets the reference score $e_{\ell_1}^{(0)}$. Each subsequent candidate is admitted only when $e_{\ell_{m+1}}^{(m)}\geq\tau e_{\ell_1}^{(0)}$; otherwise, selection terminates before admission. Therefore, $\tau$ directly specifies the minimum fraction of the first selected candidate's effective score that a later candidate must retain, thereby preventing the admission of candidates with weak adaptation evidence. Once admitted, $i_{m+1}$ is added to the selected set, the availability mask and regional counts are updated, and $\mathbf{d}^{(m)}$ and $\mathbf{e}^{(m)}$ are recomputed for the next round while the base-score vector $\bar{\mathbf{b}}$ remains fixed. In our experiments, the average number of selected patches is approximately six. The returned set is $\mathcal{S}^{t}$ with $M^{t}=m_{\star}$, where $m_{\star}$ is the number of admitted candidates.

The tensorized design greatly reduces the time cost of evidence gathering and patch selection (Table~\ref{tab:tensorized-greedy-runtime}, Appendix~\ref{app:tensorized-greedy-scan}), enabling ReTailor to select informative patches without compromising deployment efficiency.

\subsection{Rolling Residual Bootstrap}
\label{sec:rolling-bootstrap}

\begin{figure*}[t]
    \centering
    \includegraphics[width=\textwidth,trim=0 26pt 52pt 0,clip]{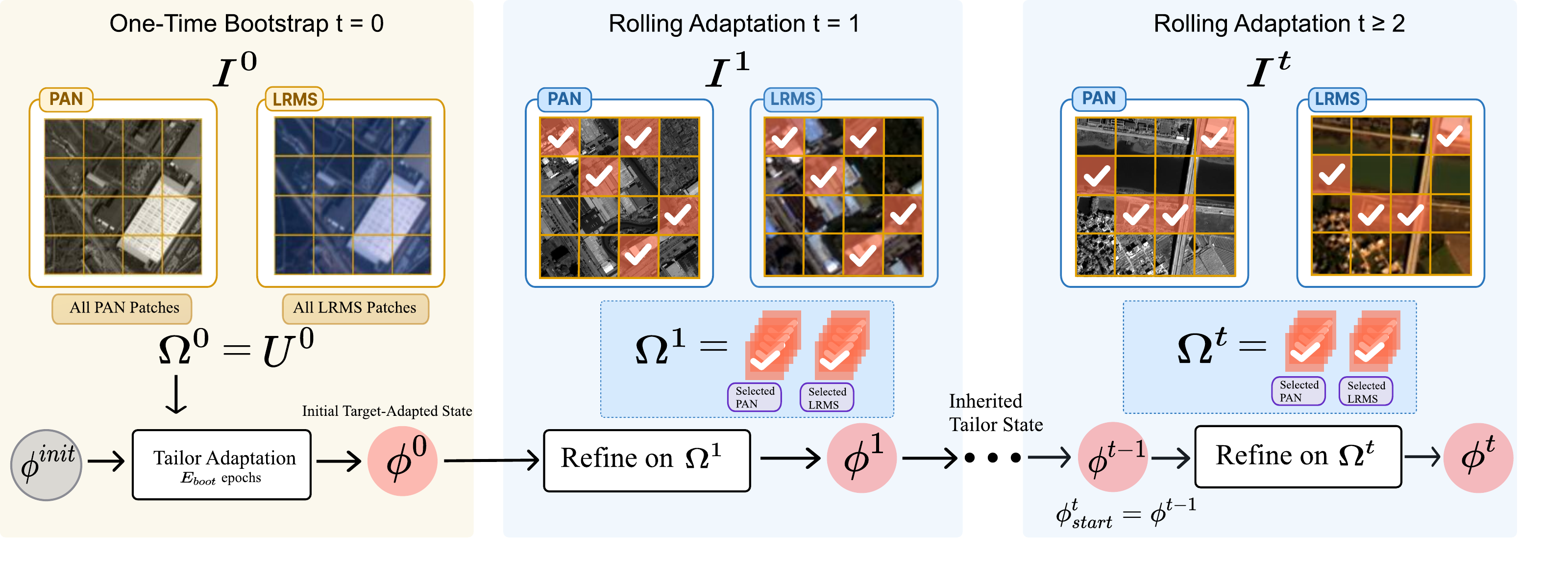}
    \caption{\textbf{Rolling residual bootstrap.} The seed image adapts $\phi_{\mathrm{init}}$ on all candidates to give $\phi^{0}$. Each later image inherits $\phi^{t-1}$, refines it on its selected patches, and passes $\phi^{t}$ on. Only Tailor parameters are transferred, and Adam is reinitialized for each image.}
    \label{fig:rolling-bootstrap}
\end{figure*}

Consider a deployment stream $\{\mathcal{I}^{t}\}_{t=0}^{T}$ of target images requiring pansharpening, all acquired by the same target sensor. Because the mismatch is dominated by sensor-dependent discrepancies, the corrections needed by different images in the stream are related: each combines a shared sensor-related component with image-specific variation. Section~\ref{sec:residual-probe} formulates and verifies this in residual space (see also Fig.~\ref{3in1}(a)). Reusing the previous Tailor parameters to initialize adaptation retains the previously learned sensor-related component, so the objective of Section~\ref{sec:adaptation-objective} only needs to focus on refining the image-specific part.

To establish this shared sensor-related prior at $t=0$, we perform full-patch adaptation across all $N_c$ candidates of the seed image from the zero-output initialization $\phi_{\mathrm{init}}$, yielding $\phi^{0}$. For each $t\geq1$, adaptation starts from the inherited $\phi^{t-1}$ and is optimized on the selected set $\Omega^{t}$, giving $\phi^{t}$, which is then used for full-patch inference on image $t$ and passed to image $t+1$ (Fig.~\ref{fig:rolling-bootstrap}). Although only the seed image undergoes full-patch adaptation while subsequent images use selected subsets, overall performance is insensitive to the choice of the seed image, as verified in Section~\ref{sec:first-image-robustness}; this observation is consistent with the shared correction component principle behind the bootstrap.

\section{Experiments and Results}
\label{sec:experiments}

Section~\ref{sec:experimental-setup} describes the setup. Section~\ref{sec:main-results} reports the main cross-sensor results, Section~\ref{sec:lora-control} compares ReTailor with parameter-matched LoRA, and Section~\ref{sec:cross-task-generalizability} transfers the Feature Tailor to two other remote-sensing tasks.

\subsection{Experimental Setup}
\label{sec:experimental-setup}

\textbf{Datasets and backbones.}
We evaluate QB$\rightarrow$GF2 and WV3$\rightarrow$WV2 on the PanCollection benchmark~\cite{deng2022benchmark}, with FusionNet~\cite{deng2021fusionnet} and U2Net~\cite{peng2023u2net} as frozen source-pretrained backbones (Appendix~\ref{app:backbone-selection}).

\textbf{Compared methods.}
The baselines comprise five traditional methods (BT-H~\cite{lolli2017haze}, C-BDSD~\cite{garzelli2014cbdso}, BDSD-PC~\cite{vivone2019robust}, MTF-GLP~\cite{aiazzi2006mtf}, MF~\cite{restaino2016morphological}), the source-pretrained FusionNet and U2Net applied directly, two learning-based methods (WFANet~\cite{huang2025wfanet}, FusionMamba~\cite{peng2024fusionmamba}), and two zero-shot methods (PsDip~\cite{rui2024variationalzs}, ZS-Pan~\cite{cao2024zspan}). FT-only is our reference variant: the frozen backbone with the Feature Tailor adapted on eight random patches and cold-started for each image (budget chosen in Appendix~\ref{app:patch-sensitivity}). It lacks only content-aware selection and rolling residual bootstrap, and thus isolates the residual design. Section~\ref{sec:lora-control} further compares with LoRA~\cite{hu2022lora}, a representative parameter-efficient fine-tuning method.

\textbf{Metrics and timing.}
We use the reference-free HQNR, $D_{\lambda}$ (spectral distortion), and $D_s$ (spatial distortion)~\mbox{\cite{aiazzi2014hqnr,alparone2008qnr}}, where higher HQNR and lower $D_{\lambda}$ and $D_s$ are better, and report image-level mean $\pm$ standard deviation. The seed image, i.e., the first image in the stream, is excluded from all reported metrics. It alone undergoes full-patch adaptation (Section~\ref{sec:rolling-bootstrap}), which may give it higher scores and thereby inflate the reported mean. In deployment, its one-time cost is amortized over the stream. For segmentation we report aAcc, mIoU, mAcc, and building IoU and accuracy, emphasizing mIoU and building IoU since the background dominates.

\subsection{Main Cross-Sensor Pansharpening Results}
\label{sec:main-results}

\begin{table*}[!t]
\centering
\caption{\textbf{Main cross-sensor results on QB$\rightarrow$GF2 and WV3$\rightarrow$WV2.} (a) Fusion quality; best in bold, second-best underlined. (b) Selected patches and adaptation time (patch selection plus Tailor optimization, excluding the identical full-patch inference); better in bold.}
\label{tab:main-results}
\standardtablefont
\setlength{\tabcolsep}{2.5pt}
\renewcommand{\arraystretch}{1.08}
\textbf{(a) Fusion quality.}\\[0.3em]
\begin{adjustbox}{max width=\textwidth}
\begin{tabular}{@{}lcccccc@{}}
\toprule
& \multicolumn{3}{c}{QB$\rightarrow$GF2} & \multicolumn{3}{c}{WV3$\rightarrow$WV2} \\
\cmidrule(lr){2-4}\cmidrule(lr){5-7}
Method & HQNR $\uparrow$ & $D_{\lambda}\downarrow$ & $D_s\downarrow$ & HQNR $\uparrow$ & $D_{\lambda}\downarrow$ & $D_s\downarrow$ \\
\midrule
BT-H & $0.8155\pm0.0288$ & $0.0603\pm0.0233$ & $0.1322\pm0.0188$ & $0.8414\pm0.0413$ & $0.0790\pm0.0329$ & $0.0868\pm0.0160$ \\
C-BDSD & $0.6996\pm0.0323$ & $0.2149\pm0.0302$ & $0.1091\pm0.0154$ & $0.6956\pm0.0461$ & $0.2253\pm0.0488$ & $0.1019\pm0.0296$ \\
BDSD-PC & $0.7797\pm0.0396$ & $0.0769\pm0.0285$ & $0.1557\pm0.0277$ & $0.8228\pm0.0344$ & $0.1439\pm0.0272$ & $0.0391\pm0.0172$ \\
MTF-GLP & $0.8064\pm0.0443$ & $\underline{0.0421\pm0.0172}$ & $0.1586\pm0.0341$ & $0.8804\pm0.0427$ & $\underline{0.0360\pm0.0166}$ & $0.0872\pm0.0310$ \\
MF & $0.8000\pm0.0438$ & $0.0526\pm0.0196$ & $0.1560\pm0.0326$ & $0.8802\pm0.0419$ & $0.0440\pm0.0191$ & $0.0797\pm0.0265$ \\
PsDip & $0.7825\pm0.0325$ & $0.1323\pm0.0286$ & $0.0981\pm0.0256$ & $0.8980\pm0.0226$ & $0.0385\pm0.0239$ & $0.0659\pm0.0158$ \\
ZS-Pan & $\underline{0.8925\pm0.0240}$ & $0.0778\pm0.0182$ & $\bm{0.0323\pm0.0159}$ & $0.9112\pm0.0336$ & $0.0476\pm0.0270$ & $0.0435\pm0.0130$ \\
FusionMamba & $0.7627\pm0.0724$ & $0.1627\pm0.0695$ & $0.0893\pm0.0352$ & $0.9022\pm0.0242$ & $0.0572\pm0.0272$ & $0.0429\pm0.0112$ \\
FusionNet & $0.8403\pm0.0505$ & $0.1127\pm0.0499$ & $0.0532\pm0.0106$ & $0.8886\pm0.0212$ & $0.0553\pm0.0268$ & $0.0591\pm0.0154$ \\
U2Net & $0.7058\pm0.0713$ & $0.2228\pm0.0637$ & $0.0926\pm0.0437$ & $0.8686\pm0.0803$ & $0.0954\pm0.0786$ & $0.0404\pm0.0097$ \\
WFANet & $0.7088\pm0.0449$ & $0.1971\pm0.0479$ & $0.1165\pm0.0379$ & $0.9142\pm0.0296$ & $0.0510\pm0.0297$ & $0.0367\pm0.0053$ \\
\midrule
FT-only$_{\mathrm{FusionNet}}$ & $0.8871\pm0.0574$ & $0.0763\pm0.0458$ & $\underline{0.0404\pm0.0179}$ & $0.9075\pm0.0259$ & $0.0463\pm0.0217$ & $0.0485\pm0.0150$ \\
FT-only$_{\mathrm{U2Net}}$ & $0.8058\pm0.0478$ & $0.1087\pm0.0402$ & $0.0959\pm0.0367$ & $0.9200\pm0.0460$ & $0.0583\pm0.0443$ & $\bm{0.0232\pm0.0051}$ \\
ReTailor$_{\mathrm{FusionNet}}$ & $\bm{0.8976\pm0.0264}$ & $\bm{0.0339\pm0.0292}$ & $0.0706\pm0.0218$ & $\bm{0.9245\pm0.0394}$ & $\bm{0.0335\pm0.0198}$ & $0.0438\pm0.0233$ \\
ReTailor$_{\mathrm{U2Net}}$ & $0.8172\pm0.0485$ & $0.0990\pm0.0410$ & $0.0931\pm0.0338$ & $\underline{0.9215\pm0.0419}$ & $0.0508\pm0.0416$ & $\underline{0.0292\pm0.0091}$ \\
\bottomrule
\end{tabular}
\end{adjustbox}\\[0.8em]
\textbf{(b) Adaptation cost.}\\[0.3em]
\begin{tabular*}{0.90\textwidth}{@{\extracolsep{\fill}}llcccc@{}}
\toprule
& & \multicolumn{2}{c}{QB$\rightarrow$GF2} & \multicolumn{2}{c}{WV3$\rightarrow$WV2} \\
\cmidrule(lr){3-4}\cmidrule(lr){5-6}
Backbone & Method & Selected $\downarrow$ & Time (s) $\downarrow$ & Selected $\downarrow$ & Time (s) $\downarrow$ \\
\midrule
\multirow{2}{*}{FusionNet} & FT-only & $8.00\pm0.00$ & $0.1488\pm0.0363$ & $8.00\pm0.00$ & $0.2131\pm0.0824$ \\
& ReTailor & $\bm{6.47\pm1.60}$ & $\bm{0.1240\pm0.0384}$ & $\bm{6.05\pm1.61}$ & $\bm{0.1729\pm0.0568}$ \\
\addlinespace[0.2em]
\multirow{2}{*}{U2Net} & FT-only & $8.00\pm0.00$ & $0.2326\pm0.0574$ & $8.00\pm0.00$ & $0.2408\pm0.0381$ \\
& ReTailor & $\bm{6.47\pm1.60}$ & $\bm{0.2024\pm0.0758}$ & $\bm{6.05\pm1.61}$ & $\bm{0.1924\pm0.0563}$ \\
\bottomrule
\end{tabular*}
\end{table*}

\begin{figure*}[!t]
\centering
\includegraphics[width=\textwidth]{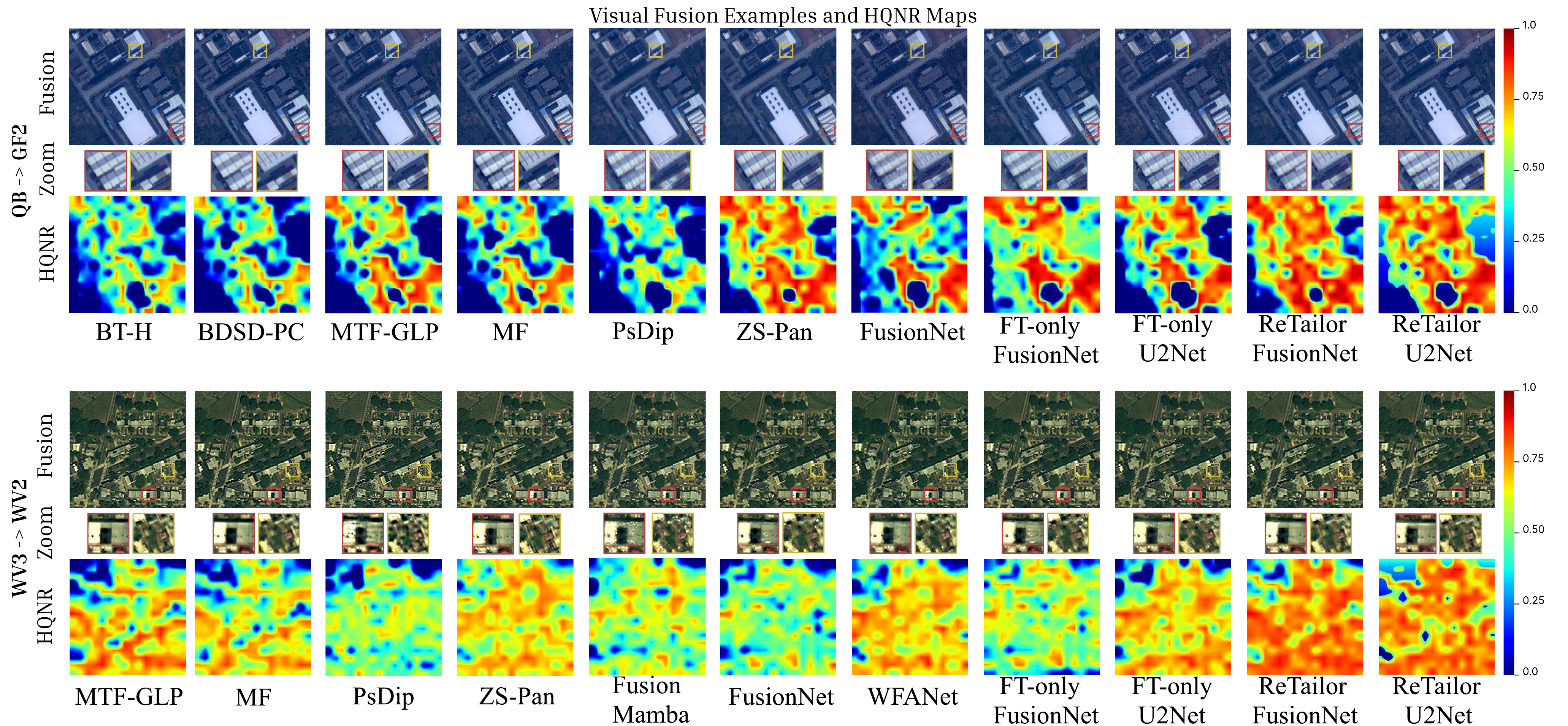}
\caption{\textbf{Qualitative comparison.} For each transfer: fused image with two regions of interest (ROIs), magnified ROIs, and HQNR map (0--1, warmer is higher). Columns: the seven baselines with the highest HQNR (order of Table~\ref{tab:main-results}) and the FT-only and ReTailor variants.}
\label{fig:qualitative-comparison}
\end{figure*}

Table~\ref{tab:main-results}(a) shows that ReTailor$_{\mathrm{FusionNet}}$ achieves the best HQNR and $D_{\lambda}$ on both transfers (0.8976 and 0.9245), exceeding the best traditional method in HQNR by 0.0821 and 0.0441. The next-best HQNR is ReTailor$_{\mathrm{U2Net}}$ on WV3$\rightarrow$WV2 and the zero-shot ZS-Pan on QB$\rightarrow$GF2. The Tailor also lifts weaker source models: for U2Net, HQNR rises from 0.7058 to 0.8172 on QB$\rightarrow$GF2 and from 0.8686 to 0.9215 on WV3$\rightarrow$WV2. Compared with FT-only, ReTailor improves HQNR in all four backbone--transfer settings with fewer patches (6.05--6.47 on average) and 13.0\%--20.1\% shorter adaptation time (Table~\ref{tab:main-results}(b)). Fig.~\ref{fig:qualitative-comparison} gives visual examples.

\subsection{Comparison with Parameter-Matched LoRA}
\label{sec:lora-control}

To test whether ReTailor's gains go beyond a small trainable-parameter count, we compare it with standard LoRA at closely matched parameter counts, using the same image stream, content-aware patches, and adaptation epochs per image. ReTailor adapts one residual module at a frozen interface and inherits it across images, whereas LoRA attaches low-rank branches to several layers and is cold-started for each image, following its standard per-image paradigm.

\begin{table*}[!t]
\centering
\caption{\textbf{ReTailor versus parameter-matched LoRA~\cite{hu2022lora}.} Both use content-aware patches and 10 adaptation epochs per image; better in bold.}
\label{tab:lora-control}
\standardtablefont
\setlength{\tabcolsep}{3pt}
\renewcommand{\arraystretch}{0.96}
\begin{adjustbox}{max width=\textwidth}
\begin{tabular}{lllcccc}
\toprule
Backbone & Transfer & Method & Params & HQNR $\uparrow$ & $D_{\lambda}\downarrow$ & $D_s\downarrow$ \\
\midrule
\multirow{4}{*}{FusionNet} & \multirow{2}{*}{WV3$\rightarrow$WV2} & ReTailor & 221,472 & \textbf{0.9245 $\pm$ 0.0394} & \textbf{0.0335 $\pm$ 0.0198} & 0.0438 $\pm$ 0.0233 \\
& & LoRA & 222,000 & 0.9241 $\pm$ 0.0297 & 0.0449 $\pm$ 0.0232 & \textbf{0.0326 $\pm$ 0.0126} \\
& \multirow{2}{*}{QB$\rightarrow$GF2} & ReTailor & 221,472 & 0.8976 $\pm$ 0.0264 & \textbf{0.0339 $\pm$ 0.0292} & 0.0706 $\pm$ 0.0218 \\
& & LoRA & 221,920 & \textbf{0.9122 $\pm$ 0.0466} & 0.0646 $\pm$ 0.0428 & \textbf{0.0250 $\pm$ 0.0079} \\
\addlinespace[0.2em]
\multirow{4}{*}{U2Net} & \multirow{2}{*}{WV3$\rightarrow$WV2} & ReTailor & 221,472 & \textbf{0.9215 $\pm$ 0.0419} & \textbf{0.0508 $\pm$ 0.0416} & \textbf{0.0292 $\pm$ 0.0091} \\
& & LoRA & 224,288 & 0.9062 $\pm$ 0.0463 & 0.0653 $\pm$ 0.0437 & 0.0307 $\pm$ 0.0087 \\
& \multirow{2}{*}{QB$\rightarrow$GF2} & ReTailor & 221,472 & \textbf{0.8172 $\pm$ 0.0485} & \textbf{0.0990 $\pm$ 0.0410} & \textbf{0.0931 $\pm$ 0.0338} \\
& & LoRA & 226,677 & 0.7326 $\pm$ 0.0625 & 0.1773 $\pm$ 0.0525 & 0.1102 $\pm$ 0.0416 \\
\bottomrule
\end{tabular}
\end{adjustbox}
\end{table*}

\begin{figure}[pos=t]
    \centering
    \includegraphics[width=0.84\columnwidth]{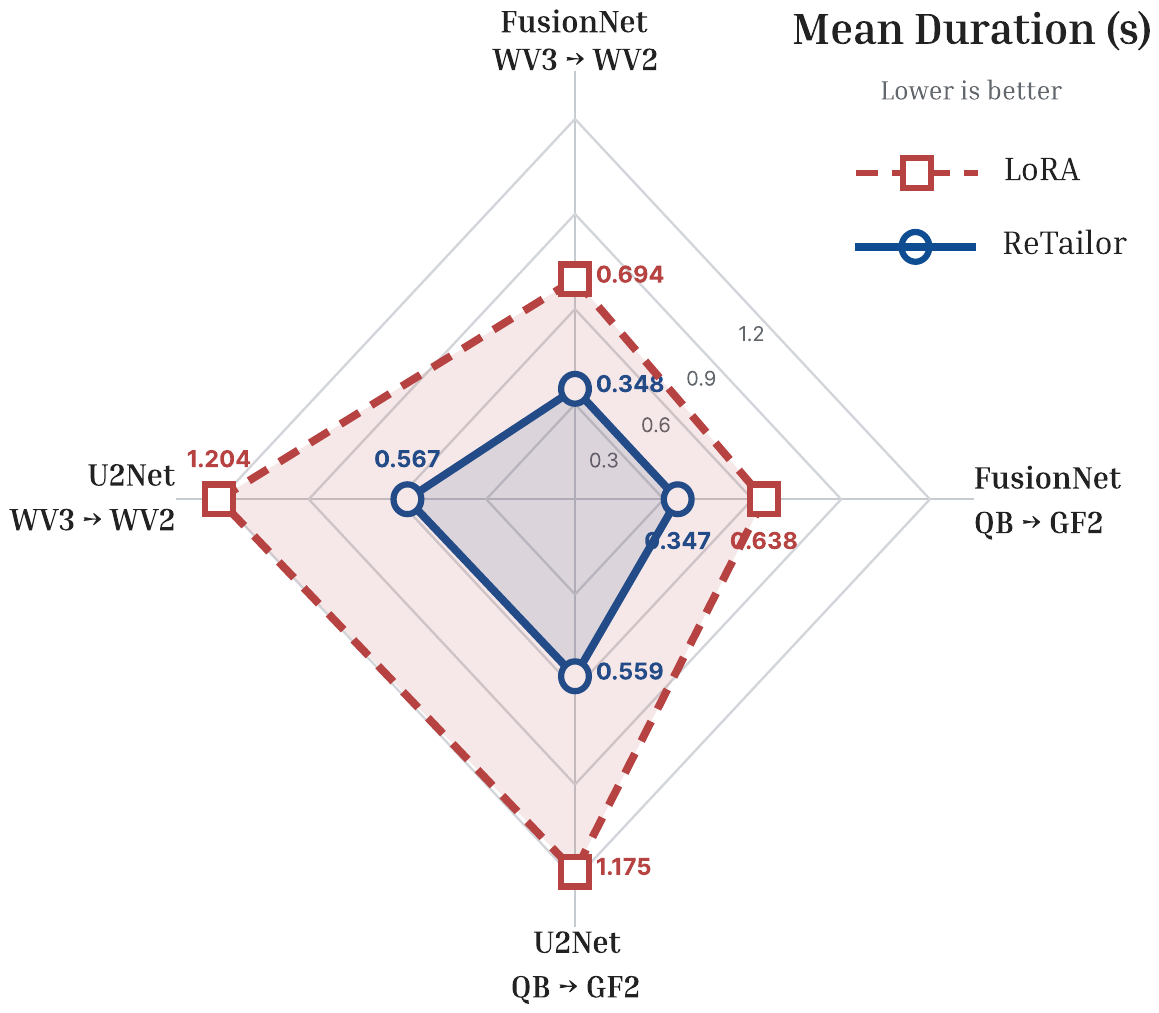}
    \caption{\textbf{End-to-end duration of ReTailor and LoRA.} Mean per-image latency in the four backbone--transfer settings (smaller radius is faster); ReTailor's latency includes patch selection, adaptation, and full-patch inference.}
    \label{fig:lora-duration-radar}
\end{figure}

Fig.~\ref{fig:lora-duration-radar} shows that ReTailor cuts mean end-to-end duration (including patch selection, adaptation, and full-patch inference; LoRA's timing excludes patch selection) by 45.6\%--52.9\% in all four settings, and Table~\ref{tab:lora-control} shows that it lowers $D_{\lambda}$ by 22.2\%--47.5\%, achieves higher HQNR in three settings, and lower $D_s$ on both U2Net transfers. At matched capacity, a compact residual correction inherited across images is thus both faster and more faithful than layer-wise low-rank updates learned from scratch for each image.

\subsection{Generalization to Other Cross-Sensor Tasks}
\label{sec:cross-task-generalizability}

We test whether the Feature Tailor, ReTailor's core update module, transfers to arbitrary-scale super-resolution pansharpening (G-ZAP) and semantic segmentation (CrossEarth). In both, the original network is frozen and only the inserted Tailor is trained (Appendix~\ref{app:cross-task-settings}).

\textbf{Super-resolution pansharpening.}
G-ZAP performs arbitrary-scale pansharpening with an implicit neural representation~\cite{yang2026gzap}. The Tailor is adapted from unlabeled target PAN/LRMS observations. Table~\ref{tab:gzap-results} shows gains on all metrics for both transfers. On QB$\rightarrow$GF2, HQNR rises from 0.8344 to 0.9508, with $D_{\lambda}$ and $D_s$ falling from 0.0793 and 0.0940 to 0.0203 and 0.0295. On WV3$\rightarrow$WV2, HQNR rises from 0.9376 to 0.9468.

\begin{table}[!t]
\centering
\caption{\textbf{Feature Tailor on G-ZAP~\cite{yang2026gzap}.} Mean $\pm$ std over 20 test images per transfer; better in bold.}
\label{tab:gzap-results}
\standardtablefont
\renewcommand{\arraystretch}{1.08}
\begin{adjustbox}{max width=\columnwidth}
\begin{tabular}{@{}lccc@{}}
\toprule
Method & HQNR $\uparrow$ & $D_{\lambda}\downarrow$ & $D_s\downarrow$ \\
\midrule
\multicolumn{4}{@{}l}{\textit{QB$\rightarrow$GF2}} \\
G-ZAP & $0.8344 \pm 0.0319$ & $0.0793 \pm 0.0232$ & $0.0940 \pm 0.0189$ \\
G-ZAP + FT & $\mathbf{0.9508 \pm 0.0143}$ & $\mathbf{0.0203 \pm 0.0076}$ & $\mathbf{0.0295 \pm 0.0132}$ \\
\addlinespace[0.2em]
\multicolumn{4}{@{}l}{\textit{WV3$\rightarrow$WV2}} \\
G-ZAP & $0.9376 \pm 0.0229$ & $0.0340 \pm 0.0210$ & $0.0294 \pm 0.0065$ \\
G-ZAP + FT & $\mathbf{0.9468 \pm 0.0129}$ & $\mathbf{0.0254 \pm 0.0096}$ & $\mathbf{0.0285 \pm 0.0091}$ \\
\bottomrule
\end{tabular}
\end{adjustbox}
\end{table}

\begin{figure*}[!t]
    \centering
    \includegraphics[width=\textwidth]{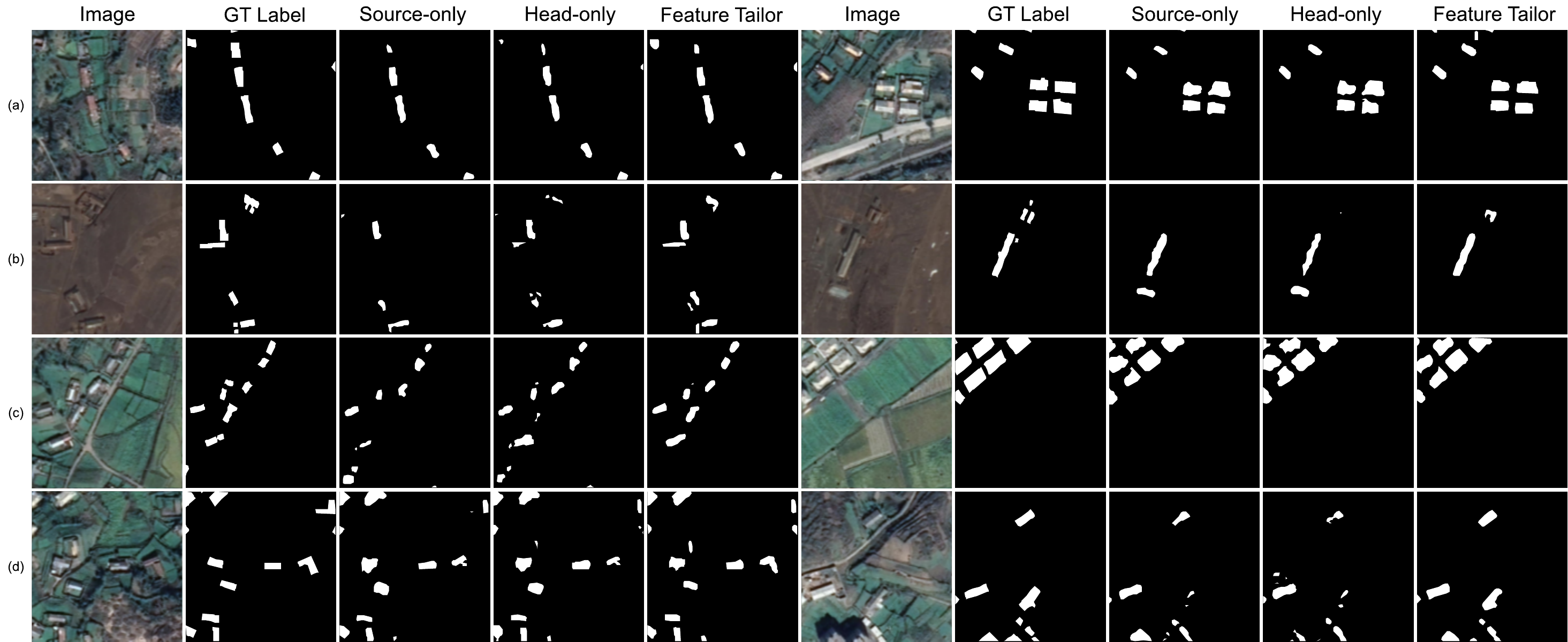}
    \caption{\textbf{Qualitative CrossEarth segmentation~\mbox{\cite{gong2026crossearth,ji2019whu}}.} First eight Satellite test samples; columns: input, ground truth, Source-only, Head-only, Feature Tailor (white: building).}
    \label{fig:crossearth-qualitative}
\end{figure*}

\textbf{Semantic segmentation.}
On the CrossEarth Aerial-to-Satellite protocol, labeled target images are used to predict building masks. We compare Source-only with two target-supervised variants under the same schedule: Head-only, which updates the original output head (14.47M parameters), and Feature Tailor (14.16M parameters, 2.18\% fewer). In Table~\ref{tab:crossearth-results}, Feature Tailor improves every metric, raising mIoU from 74.74 to 85.09 and building IoU from 50.26 to 70.61 over Head-only. The larger gain on building metrics than on aAcc shows that the improvement is not driven by the dominant background class, and Fig.~\ref{fig:crossearth-qualitative} shows more complete building regions, including small and fragmented ones.

\begin{table}[!t]
\centering
\caption{\textbf{CrossEarth Aerial-to-Satellite segmentation~\mbox{\cite{gong2026crossearth,ji2019whu}} (\%).} Best in bold, second-best underlined.}
\label{tab:crossearth-results}
\standardtablefont
\setlength{\tabcolsep}{1.8pt}
\renewcommand{\arraystretch}{1.08}
\begin{tabular*}{\columnwidth}{@{\extracolsep{\fill}}lccccc@{}}
\toprule
Method & aAcc $\uparrow$ & mIoU $\uparrow$ & mAcc $\uparrow$ & \shortstack{Bldg.\\IoU $\uparrow$} & \shortstack{Bldg.\\Acc. $\uparrow$} \\
\midrule
Source-only & 99.10 & 71.67 & 78.20 & 44.25 & 56.76 \\
Head-only & \underline{99.23} & \underline{74.74} & \underline{80.64} & \underline{50.26} & \underline{61.56} \\
Feature Tailor & \textbf{99.57} & \textbf{85.09} & \textbf{90.97} & \textbf{70.61} & \textbf{82.14} \\
\bottomrule
\end{tabular*}
\end{table}

\section{Ablation Study}
\label{sec:ablation}

We ablate ReTailor's two complementary components one at a time: content-aware patch selection retains the most informative patches to improve fusion quality, and rolling residual bootstrap preserves the learned sensor-related correction across images. Section~\ref{sec:selector-ablation} compares selection with random sampling and examines its budget, score composition, regional discounting, and scalability. Section~\ref{sec:rolling-ablation} compares rolling with cold start and examines whether corrections are shared across images and whether the stream is robust to its seed image and length. All variants share the same backbone, 10-epoch schedule, and optimizer, with FT-only (Random-8, cold start) as the baseline.

\subsection{Content-Aware Patch Selection}
\label{sec:selector-ablation}

Selection acts within each image. Section~\ref{sec:selection-quality} tests whether content-aware scoring picks better patches than random sampling and whether adaptive stopping reduces cost, Section~\ref{sec:evidence-composition} justifies anchoring the score on PAN variation, Section~\ref{sec:regional-discount} verifies the contribution of regional discounting, and Section~\ref{sec:large-pool} tests whether the advantage persists on large images.

\subsubsection{Selection quality and budget}
\label{sec:selection-quality}
Random-8 and Content-aware-8 share one budget, so their difference isolates which patches are chosen. Content-aware-Adaptive adds relative stopping to the same scoring, isolating the effect of a smaller, image-dependent budget.

\begin{table*}[!t]
\centering
\caption{\textbf{Ablation of patch selection.} All variants use per-image cold start; best in bold, second-best underlined within each backbone--transfer block.}
\label{tab:selector-ablation}
\standardtablefont
\begin{adjustbox}{max width=\textwidth}
\setlength{\tabcolsep}{3pt}
\begin{tabular*}{\textwidth}{@{\extracolsep{\fill}}lllccc@{}}
\toprule
Backbone & Transfer & Selector & HQNR $\uparrow$ & $D_{\lambda}\downarrow$ & $D_s\downarrow$ \\
\midrule
\multirow{6}{*}{FusionNet} & \multirow{3}{*}{WV3$\rightarrow$WV2} & Random-8 & $0.9075\pm0.0259$ & $0.0463\pm0.0217$ & $0.0485\pm0.0150$ \\
& & Content-aware-8 & $\bm{0.9186\pm0.0261}$ & $\bm{0.0385\pm0.0241}$ & $\bm{0.0446\pm0.0128}$ \\
& & Content-aware-Adaptive & $\underline{0.9088\pm0.0224}$ & $\underline{0.0453\pm0.0211}$ & $\underline{0.0481\pm0.0146}$ \\
\addlinespace[0.35em]
& \multirow{3}{*}{QB$\rightarrow$GF2} & Random-8 & $0.8871\pm0.0574$ & $0.0763\pm0.0458$ & $\bm{0.0404\pm0.0179}$ \\
& & Content-aware-8 & $\bm{0.8977\pm0.0411}$ & $\bm{0.0612\pm0.0408}$ & $0.0438\pm0.0138$ \\
& & Content-aware-Adaptive & $\underline{0.8901\pm0.0467}$ & $\underline{0.0718\pm0.0478}$ & $\underline{0.0411\pm0.0142}$ \\
\addlinespace[0.55em]
\multirow{6}{*}{U2Net} & \multirow{3}{*}{WV3$\rightarrow$WV2} & Random-8 & $0.9200\pm0.0460$ & $0.0583\pm0.0443$ & $\bm{0.0232\pm0.0051}$ \\
& & Content-aware-8 & $\bm{0.9265\pm0.0442}$ & $\underline{0.0512\pm0.0450}$ & $\underline{0.0235\pm0.0052}$ \\
& & Content-aware-Adaptive & $\underline{0.9217\pm0.0193}$ & $\bm{0.0418\pm0.0156}$ & $0.0381\pm0.0248$ \\
\addlinespace[0.35em]
& \multirow{3}{*}{QB$\rightarrow$GF2} & Random-8 & $0.8058\pm0.0478$ & $0.1087\pm0.0402$ & $0.0959\pm0.0367$ \\
& & Content-aware-8 & $\bm{0.8180\pm0.0539}$ & $\bm{0.0980\pm0.0485}$ & $\underline{0.0931\pm0.0346}$ \\
& & Content-aware-Adaptive & $\underline{0.8136\pm0.0274}$ & $\underline{0.1073\pm0.0187}$ & $\bm{0.0886\pm0.0231}$ \\
\bottomrule
\end{tabular*}
\end{adjustbox}
\end{table*}

\begin{figure}[pos=t]
    \centering
    \includegraphics[width=\columnwidth]{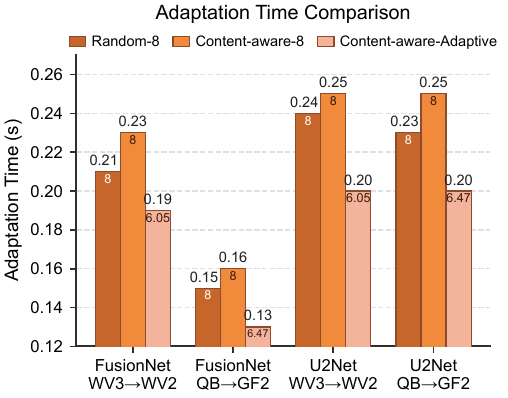}
    \caption{\textbf{Adaptation time of the three selectors.} Bars: mean per-image time (patch selection plus Tailor optimization); numbers inside: mean selected patches (Table~\ref{tab:selector-ablation}). The vertical axis starts at 0.12~s.}
    \label{fig:selector-adaptation-time}
\end{figure}

In Table~\ref{tab:selector-ablation}, Content-aware-8 gives higher HQNR and lower $D_{\lambda}$ than Random-8 in all four settings, so observation-based scoring supplies better adaptation evidence at the same budget. Content-aware-Adaptive retains these advantages over Random-8 in all four settings while cutting adaptation time by 9.9\%--17.1\% (Fig.~\ref{fig:selector-adaptation-time}).

\subsubsection{Evidence composition}
\label{sec:evidence-composition}
The selector anchors its score on PAN variation and adds LRMS inter-band variation as a supplement (Eq.~\eqref{eq:base-utility}). To justify this asymmetry, we compare PAN-only-8, LRMS-only-8, LRMS-anchored-8 (roles reversed, same $\lambda_c=0.2$), and Content-aware-8 using FusionNet on QB$\rightarrow$GF2, with per-image cold start as in Table~\ref{tab:selector-ablation}.

\begin{table}[!tb]
\centering
\caption{\textbf{Ablation of evidence composition.} QB$\rightarrow$GF2, FusionNet, eight patches per variant, per-image cold start; best in bold, second-best underlined.}
\label{tab:pan-anchor-ablation}
\standardtablefont
\setlength{\tabcolsep}{3pt}
\begin{tabular*}{\columnwidth}{@{\extracolsep{\fill}}lccc@{}}
    \toprule
    Variant & HQNR $\uparrow$ & $D_{\lambda}\downarrow$ & $D_s\downarrow$ \\
    \midrule
    PAN-only-8 & $\underline{0.8794\pm0.0502}$ & $\underline{0.0824\pm0.0409}$ & $\bm{0.0422\pm0.0158}$ \\
    LRMS-only-8 & $0.8675\pm0.0629$ & $0.0910\pm0.0504$ & $0.0466\pm0.0201$ \\
    LRMS-anchored-8 & $0.8688\pm0.0590$ & $0.0896\pm0.0469$ & $0.0466\pm0.0191$ \\
    Content-aware-8 & $\bm{0.8977\pm0.0411}$ & $\bm{0.0612\pm0.0408}$ & $\underline{0.0438\pm0.0138}$ \\
    \bottomrule
\end{tabular*}
\end{table}

PAN-only-8 outperforms LRMS-only-8 (HQNR +0.0119), so PAN variation is the stronger cue. Reversing the roles barely helps (HQNR 0.8675 to 0.8688), whereas the PAN-anchored Content-aware-8 reaches the best HQNR (0.8977) and $D_{\lambda}$ (0.0612). PAN variation should thus anchor the score, with LRMS variation adding complementary spectral evidence.

\subsubsection{Regional discount}
\label{sec:regional-discount}
Regional discounting lowers the priority of candidates whose coarse region
is already represented (Sec.~\ref{sec:content-aware-selection}). To verify
this design, we compare Random-8, No-discount-8 ($\lambda_r=0$, which
admits the eight highest-scoring candidates regardless of their regions),
and Content-aware-8 using FusionNet on both transfers, with per-image cold
start as in Table~\ref{tab:selector-ablation}.

\begin{table}[!tb]
\centering
\caption{\textbf{Ablation of regional discounting.} FusionNet, eight patches
per variant, per-image cold start; best in bold, second-best underlined
within each transfer block. Random-8 and Content-aware-8 replicate
Table~\ref{tab:selector-ablation}.}
\label{tab:regional-discount-ablation}
\standardtablefont
\setlength{\tabcolsep}{3pt}
\begin{tabular*}{\columnwidth}{@{\extracolsep{\fill}}lccc@{}}
    \toprule
    Variant & HQNR $\uparrow$ & $D_{\lambda}\downarrow$ & $D_s\downarrow$ \\
    \midrule
    \multicolumn{4}{l}{\emph{WV3$\rightarrow$WV2}} \\
    Random-8 & $0.9075\pm0.0259$ & $0.0463\pm0.0217$ & $\underline{0.0485\pm0.0150}$ \\
    No-discount-8 & $\underline{0.9083\pm0.0236}$ & $\underline{0.0428\pm0.0235}$ & $0.0512\pm0.0140$ \\
    Content-aware-8 & $\bm{0.9186\pm0.0261}$ & $\bm{0.0385\pm0.0241}$ & $\bm{0.0446\pm0.0128}$ \\
    \addlinespace[0.2em]
    \multicolumn{4}{l}{\emph{QB$\rightarrow$GF2}} \\
    Random-8 & $0.8871\pm0.0574$ & $0.0763\pm0.0458$ & $\bm{0.0404\pm0.0179}$ \\
    No-discount-8 & $\underline{0.8916\pm0.0499}$ & $\underline{0.0684\pm0.0438}$ & $\underline{0.0431\pm0.0129}$ \\
    Content-aware-8 & $\bm{0.8977\pm0.0411}$ & $\bm{0.0612\pm0.0408}$ & $0.0438\pm0.0138$ \\
    \bottomrule
\end{tabular*}
\end{table}

In Table~\ref{tab:regional-discount-ablation}, No-discount-8 outperforms
Random-8 on both transfers (HQNR +0.0008 and +0.0045; $D_{\lambda}$ down to
0.0428 and 0.0684), so the base score itself identifies informative
patches. Adding regional discounting further raises HQNR by 0.0103 and
0.0061 (to 0.9186 and 0.8977) and lowers $D_{\lambda}$ to 0.0385 and
0.0612. The score and the discount are therefore complementary: the former
locates informative patches, and the latter keeps the fixed budget from
clustering in a few high-variation regions.

\subsubsection{Large images}
\label{sec:large-pool}
Intuitively, enlarging the image hurts random selection but helps content-aware selection: as the candidate pool grows, random sampling is more likely to admit weak patches for Tailor adaptation, while the larger image also produces more good and more diverse patches for the scorer to pick from. Content-aware selection should therefore widen its lead over random as the image grows. To verify this, we compare Random-8 with Content-aware-8 on two $4{,}000\times4{,}000$ images, one per transfer, with per-image cold start as in Table~\ref{tab:selector-ablation}.

\begin{table*}[!t]
    \centering
    \caption{\textbf{Patch selection on large images.} One $4{,}000\times4{,}000$ image each from WV2 and GF2, eight patches per method, per-image cold start; better in bold.}
    \label{tab:selector-composite}
    \standardtablefont
    \begin{tabular*}{0.95\textwidth}{@{\extracolsep{\fill}}lllccc@{}}
        \toprule
        Backbone & Transfer & Selection & HQNR $\uparrow$ & $D_{\lambda}\downarrow$ & $D_s\downarrow$ \\
        \midrule
        \multirow{4}{*}{FusionNet}
        & \multirow{2}{*}{WV3$\rightarrow$WV2}
        & Random-8 & 0.9266 & 0.0579 & 0.0165 \\
        & & Content-aware-8 & \textbf{0.9513} & \textbf{0.0392} & \textbf{0.0098} \\
        & \multirow{2}{*}{QB$\rightarrow$GF2}
        & Random-8 & 0.8211 & 0.1378 & 0.0476 \\
        & & Content-aware-8 & \textbf{0.8751} & \textbf{0.1012} & \textbf{0.0263} \\
        \addlinespace[0.2em]
        \multirow{4}{*}{U2Net}
        & \multirow{2}{*}{WV3$\rightarrow$WV2}
        & Random-8 & 0.8358 & 0.1049 & 0.0663 \\
        & & Content-aware-8 & \textbf{0.8808} & \textbf{0.0887} & \textbf{0.0334} \\
        & \multirow{2}{*}{QB$\rightarrow$GF2}
        & Random-8 & 0.7199 & 0.1982 & 0.1021 \\
        & & Content-aware-8 & \textbf{0.7559} & \textbf{0.1684} & \textbf{0.0910} \\
        \bottomrule
    \end{tabular*}
\end{table*}

In Table~\ref{tab:selector-composite}, content-aware selection remains better than random selection on all three metrics in all four cases, and its HQNR gains widen to 0.0247--0.0540 from 0.0065--0.0122 on standard images (Table~\ref{tab:selector-ablation}), consistent with the intuition above: random sampling admits more uninformative patches from the larger pool, whereas the scorer benefits from the richer and more diverse informative regions that the larger image supplies.

\subsection{Rolling Residual Bootstrap}
\label{sec:rolling-ablation}

Rolling residual bootstrap acts along the image stream. Section~\ref{sec:residual-probe} probes whether the reused corrections are shared across images, Section~\ref{sec:rolling-vs-cold} compares rolling with per-image cold start under matched budgets, and Sections~\ref{sec:first-image-robustness} and~\ref{sec:long-sequence-stability} further test the design by varying the seed image and lengthening the stream.

\begin{table*}[!t]
\centering
\caption{\textbf{Ablation of rolling residual bootstrap.} Both arms use Random-8; better in bold. Time covers patch selection and Tailor optimization.}
\label{tab:rolling-ablation}
\standardtablefont
\renewcommand{\arraystretch}{0.90}
\begin{tabular*}{0.95\textwidth}{@{\extracolsep{\fill}}lllcccc@{}}
\toprule
Backbone & Transfer & Initialization & HQNR $\uparrow$ & $D_{\lambda}\downarrow$ & $D_s\downarrow$ & Adaptation time (s) $\downarrow$ \\
\midrule
\multirow{4}{*}{FusionNet} & \multirow{2}{*}{WV3$\rightarrow$WV2} & Cold-start & 0.9075 $\pm$ 0.0259 & 0.0463 $\pm$ 0.0217 & 0.0485 $\pm$ 0.0150 & 0.2131 $\pm$ 0.0824 \\
& & Rolling & \textbf{0.9249 $\pm$ 0.0417} & \textbf{0.0317 $\pm$ 0.0192} & \textbf{0.0452 $\pm$ 0.0266} & \textbf{0.1974 $\pm$ 0.0581} \\
\addlinespace[0.4em]
& \multirow{2}{*}{QB$\rightarrow$GF2} & Cold-start & 0.8871 $\pm$ 0.0574 & 0.0763 $\pm$ 0.0458 & \textbf{0.0404 $\pm$ 0.0179} & 0.1488 $\pm$ 0.0363 \\
& & Rolling & \textbf{0.9074 $\pm$ 0.0212} & \textbf{0.0260 $\pm$ 0.0212} & 0.0683 $\pm$ 0.0186 & \textbf{0.1315 $\pm$ 0.0447} \\
\addlinespace[0.6em]
\multirow{4}{*}{U2Net} & \multirow{2}{*}{WV3$\rightarrow$WV2} & Cold-start & 0.9200 $\pm$ 0.0460 & 0.0583 $\pm$ 0.0443 & 0.0232 $\pm$ 0.0051 & 0.2408 $\pm$ 0.0381 \\
& & Rolling & \textbf{0.9249 $\pm$ 0.0408} & \textbf{0.0542 $\pm$ 0.0415} & \textbf{0.0221 $\pm$ 0.0045} & \textbf{0.2327 $\pm$ 0.0195} \\
\addlinespace[0.4em]
& \multirow{2}{*}{QB$\rightarrow$GF2} & Cold-start & 0.8058 $\pm$ 0.0478 & 0.1087 $\pm$ 0.0402 & 0.0959 $\pm$ 0.0367 & 0.2326 $\pm$ 0.0574 \\
& & Rolling & \textbf{0.8174 $\pm$ 0.0349} & \textbf{0.0973 $\pm$ 0.0305} & \textbf{0.0945 $\pm$ 0.0237} & \textbf{0.2194 $\pm$ 0.0283} \\
\bottomrule
\end{tabular*}
\end{table*}

\subsubsection{Shared correction across images}
\label{sec:residual-probe}
Rolling relies on the corrections required by images of one transfer being related. We formalize this as a decomposition. Let $\boldsymbol{\Delta}_{t,i}=\mathbf{Z}^{*}_{t,i}-\mathbf{Z}_{t,i}$ be the residual correction the Tailor applies to candidate patch $i$ of image $t$, and let
\begin{equation*}
    \mathbf{r}_{t}=\frac{1}{N_c}\sum_{i=1}^{N_c}\operatorname{GAP}(\boldsymbol{\Delta}_{t,i})
\end{equation*}
be its image-level embedding, where $\operatorname{GAP}(\cdot)$ is spatial global average pooling. We model
\begin{equation}
    \mathbf{r}_{t}=\mathbf{c}+\mathbf{e}_{t},
    \label{eq:shared-decomposition}
\end{equation}
where $\mathbf{c}$ is a sensor-related component determined by the transfer (source model, target sensor, and interface) and common to all images, and $\mathbf{e}_{t}$ is image-specific variation caused by scene content and noise. If $\mathbf{e}_{t}$ has zero mean and is uncorrelated across images, then $\mathbb{E}[\mathbf{r}_{a}^{\top}\mathbf{r}_{b}]=\|\mathbf{c}\|_2^{2}$ and $\mathbb{E}\|\mathbf{r}_{t}\|_2^{2}=\|\mathbf{c}\|_2^{2}+\mathbb{E}\|\mathbf{e}_{t}\|_2^{2}$ for $a\neq b$, so the mean pairwise cosine similarity of $\mathbf{r}_t$ approximates
\begin{equation*}
    \rho=\frac{\|\mathbf{c}\|_2^{2}}{\|\mathbf{c}\|_2^{2}+\mathbb{E}\|\mathbf{e}_{t}\|_2^{2}}\in[0,1],
\end{equation*}
the share of the correction that is sensor-related. Eq.~\eqref{eq:shared-decomposition} thus has two testable implications: $\rho$ should be high, and the residual similarity should exceed the similarity of the frozen features $\mathbf{z}_t$ (defined like $\mathbf{r}_t$ from $\mathbf{Z}_{t,i}$), which serves as a control for similarity unrelated to the correction. A high $\rho$ also explains why rolling helps: the inherited Tailor already contains $\mathbf{c}$, so adaptation only needs to learn $\mathbf{e}_{t}$, which Section~\ref{sec:rolling-vs-cold} verifies.

To test this, we use FusionNet on both transfers. For each of the 20 target images, a Tailor is cold-started and adapted with Random-8 for 10 epochs, carrying nothing across images, so no common component is injected by inheritance. We then compute the cosine similarity over all image pairs of a transfer for $\mathbf{r}_t$ and for $\mathbf{z}_t$. The probe is analysis-only and not part of deployment.

\begin{table}[!t]
\centering
\caption{\textbf{Cosine similarity between target images.} Mean $\pm$ std over 190 image pairs per transfer (380 pooled) with independently cold-started Tailors; higher in bold.}
\label{tab:residual-probe}
\standardtablefont
\setlength{\tabcolsep}{2.5pt}
\renewcommand{\arraystretch}{1.05}
\begin{tabularx}{\columnwidth}{@{}>{\raggedright\arraybackslash}Xcc@{}}
\toprule
Transfer & Latent similarity & Residual similarity \\
\midrule
WV3$\rightarrow$WV2 & 0.6782 $\pm$ 0.2995 & \textbf{0.8537 $\pm$ 0.1107} \\
QB$\rightarrow$GF2 & 0.8684 $\pm$ 0.1089 & \textbf{0.9383 $\pm$ 0.0427} \\
Aggregated & 0.7733 $\pm$ 0.2446 & \textbf{0.8960 $\pm$ 0.0940} \\
\bottomrule
\end{tabularx}
\end{table}

Table~\ref{tab:residual-probe} and Fig.~\ref{3in1}(a) support both implications. The residual similarity, which estimates $\rho$, is 0.8537 on WV3$\rightarrow$WV2 and 0.9383 on QB$\rightarrow$GF2, so most of the correction is sensor-related, and Fig.~\ref{3in1}(a) shows uniformly high similarity across all image pairs. It exceeds the latent similarity by 0.1755 and 0.0699 with lower variability, so the shared structure comes from the correction rather than from the features themselves. The lower $\rho$ on WV3$\rightarrow$WV2 indicates a larger image-specific part $\mathbf{e}_t$, which the per-image refinement of rolling residual bootstrap handles. Independently adapted Tailors thus converge to a common correction $\mathbf{c}$, consistent with the gain of inherited initialization in Section~\ref{sec:rolling-vs-cold}.

\subsubsection{Cold start versus rolling start}
\label{sec:rolling-vs-cold}
Both arms use Random-8, the same 10-epoch budget, and freshly created Adam optimizers; only the initial Tailor parameters differ (zero-output for cold start, the previous image's for rolling).

In Table~\ref{tab:rolling-ablation}, rolling improves HQNR and $D_{\lambda}$ in all four settings and $D_s$ in three, with no increase in adaptation time. The gain comes purely from the inherited sensor-related knowledge.

\subsubsection{Robustness to the seed image}
\label{sec:first-image-robustness}
Since $\phi^{0}$ is trained on the seed image alone, we test whether its choice matters: for each transfer, twenty sequences let every test image serve once as the seed image, with the remaining images randomly permuted and the pretrained backbone reloaded per sequence. Table~\ref{tab:first-image-robustness} shows negligible variation across sequences, so results do not hinge on the seed image, consistent with the shared correction measured above.

\begin{table}[!t]
\centering
\caption{\textbf{Seed-image resampling.} Twenty sequences per transfer, each with a different test image as seed and the others randomly permuted; the backbone is reloaded for each sequence.}
\label{tab:first-image-robustness}
\standardtablefont
\setlength{\tabcolsep}{2.5pt}
\renewcommand{\arraystretch}{1.05}
\begin{tabular*}{\columnwidth}{@{\extracolsep{\fill}}lcccc@{}}
\toprule
Transfer & Mean HQNR $\uparrow$ & Var. ($10^{-6}$) & Std & Min--Max \\
\midrule
QB$\rightarrow$GF2 & 0.8999 & 7.50 & 0.0027 & 0.8959--0.9072 \\
WV3$\rightarrow$WV2 & 0.9278 & 2.60 & 0.0016 & 0.9255--0.9312 \\
\bottomrule
\end{tabular*}
\end{table}

\subsubsection{Stability over long sequences}
\label{sec:long-sequence-stability}
Deployment streams can be much longer than the twenty-image test stream, so we extend the sequence to 100 images by resampling the target images and run ReTailor end to end. Table~\ref{tab:long-sequence-stability} shows that all metrics are essentially unchanged from the 20-image stream, indicating that the inherited correction does not drift over long sequences.

\begin{table}[!t]
\centering
\caption{\textbf{Long-sequence stability (FusionNet).} The standard 20-image stream versus 100-image sequences resampled from the same images.}
\label{tab:long-sequence-stability}
\standardtablefont
\setlength{\tabcolsep}{2.5pt}
\renewcommand{\arraystretch}{1.05}
\begin{tabular*}{\columnwidth}{@{\extracolsep{\fill}}llccc@{}}
\toprule
Transfer & Length & HQNR $\uparrow$ & $D_{\lambda}\downarrow$ & $D_s\downarrow$ \\
\midrule
\multirow{2}{*}{QB$\rightarrow$GF2} & 20 & $0.8976\pm0.0264$ & $0.0339\pm0.0292$ & $0.0706\pm0.0218$ \\
& 100 & $0.8978\pm0.0253$ & $0.0305\pm0.0279$ & $0.0736\pm0.0226$ \\
\midrule
\multirow{2}{*}{WV3$\rightarrow$WV2} & 20 & $0.9245\pm0.0394$ & $0.0335\pm0.0198$ & $0.0438\pm0.0233$ \\
& 100 & $0.9238\pm0.0391$ & $0.0328\pm0.0215$ & $0.0452\pm0.0222$ \\
\bottomrule
\end{tabular*}
\end{table}

\section{Discussion}
\label{sec:discussion}

\subsection{Why ReTailor Works}

ReTailor inserts a lightweight learnable module into a frozen pretrained fusion network, preserves the knowledge learned for the source sensor, and repairs the cross-sensor shift using only the most valuable patches; passing the learned correction from one image to the next further refines the result. Together, these components yield efficient, fast, and high-quality cross-sensor correction. Its effectiveness rests on three reasons.

\textbf{Evidence within an image.} The information needed to diagnose the sensor shift is already contained in the target PAN and LRMS observations, and a scoring rule that combines patch-level measures with the regional distribution can identify the most valuable patches quickly. These high-value patches provide sufficient evidence to support fast optimization for the new sensor, which improves ReTailor's adaptation quality at negligible cost.

\textbf{Correction at the fusion interface.} The Feature Tailor performs a concentrated, rapid correction at the interface between the frozen Feature Extractor and the frozen Channel Mapper, where a small additive residual suffices to absorb the sensor-specific component of the fused representation. This preserves the basic fusion information established during prior training while repairing the cross-sensor shift, and it corrects more efficiently than a parameter-matched LoRA, enabling fast transfer.

\textbf{Inheritance across images.} For the same cross-sensor transfer task, a considerable part of the required correction is shared across images, since the sensor response remains stable within a stream. By inheriting what the preceding images have already learned and then rapidly learning only the image-specific remainder on each new image, every image becomes a refinement of a common correction rather than an independent restart, which makes cross-sensor transfer work better along the stream.

\subsection{Beyond Pansharpening}
\label{sec:beyond-pansharpening}

Inserting a concentrated correction at a single interface of a frozen network is not specific to pansharpening: the same Feature Tailor improves an implicit-representation pansharpening network under observation-driven losses and a supervised building-segmentation network under labeled losses, with the original models kept frozen. Cross-sensor and cross-domain shifts are pervasive in remote sensing, so this single-interface correction is also worth considering for tasks beyond pansharpening where a pretrained model must be adapted cheaply.

\subsection{Applicability and Future Directions}

ReTailor targets streams from a stable target sensor with fixed MTF priors, a common case in operational pipelines, where the inherited Tailor gives each image a reusable starting point to refine. The current design still learns one Tailor per sensor pair, so a new transfer direction must restart the bootstrap. A natural next step is one Tailor that corrects the degradations of multiple sensors, sharing the one-time adaptation cost across sensors.

\section{Conclusion}
\label{sec:conclusion}

We present ReTailor, a real-time sequential framework for cross-sensor pansharpening that adapts frozen backbones without target ground truth. A lightweight residual Feature Tailor corrects cross-sensor shifts while preserving the pretrained fusion prior, content-aware patch selection supplies informative adaptation patches at negligible cost, and rolling residual bootstrap carries the sensor-related knowledge along the stream to improve quality. On QB$\rightarrow$GF2 and WV3$\rightarrow$WV2, ReTailor reaches HQNR of 0.8976 and 0.9245 with a fraction of a second of adaptation per image, and at matched parameter counts it roughly halves the latency of LoRA while lowering spectral distortion. Systematic ablations confirm the effectiveness of each component. The core principle of ReTailor also transfers to super-resolution pansharpening and semantic segmentation. These results make ReTailor a practical choice for cross-sensor pansharpening, while providing useful insights for other cross-sensor remote-sensing applications.

\section*{Acknowledgment}

This work was supported by the Project of the Department of Science and Technology of Sichuan Province under Grant 2025YFNH0001.

\noindent\textbf{CRediT authorship contribution statement.} \textbf{Jing-Ze Liang}: Conceptualization, Methodology, Software, Validation, Investigation, Writing -- original draft, Writing -- review \& editing. \textbf{Wei-Jie Xiao}: Methodology, Validation, Investigation, Writing -- review \& editing. \textbf{Zi-Han Cao}: Validation, Investigation, Visualization, Writing -- review \& editing. \textbf{Yang-Dong Li}: Investigation, Data curation, Writing -- review \& editing. \textbf{Tian-Yi Hu}: Investigation, Data curation, Writing -- review \& editing. \textbf{Liang-Jian Deng}: Supervision, Project administration, Writing -- review \& editing.

\noindent\textbf{Declaration of competing interest.} The authors declare that they have no known competing financial interests or personal relationships that could have appeared to influence the work reported in this paper.

\FloatBarrier
\appendix
\makeatletter\let\@addtocurcol\kernel@addtocurcol\makeatother
\nobalance

\section{Implementation Details}
\label{app:implementation-details}

\subsection{Pansharpening Protocol}
\label{app:main-task-settings}

All Feature Tailors use the three-layer residual topology of Section~\ref{sec:residual-feature-tailoring}; Table~\ref{tab:retailor-configurations} lists their interfaces and sizes, and Table~\ref{tab:training-settings} lists the optimization settings. The LeakyReLU slope is $0.2$ for G-ZAP and $0.01$ otherwise. Within each comparison only the studied factor changes, and timing is compared only between arms run on the same GPU (RTX 4090).

Each target image is a $512\times512$ PAN image with its $128\times128$ LRMS counterpart, partitioned into an $8\times8$ grid of $N_c=64$ non-overlapping $64\times64$ candidate patches; a four-pixel context rim is used at inference. Batch sizes are 8 for adaptation and 32 for inference. Random selection and test-image sampling use a fixed seed. Random selection uses $M=8$, and content-aware selection uses a $2\times2$ region grid with $\lambda_c=0.2$, $\lambda_r=0.1$, and $\tau=0.55$.

For both backbones, $\mathcal{D}_{\mathrm{MS}}^{(b)}$ applies the MS MTF kernel followed by fourfold nearest-neighbor downsampling. The MTF kernels use the per-sensor GNyq values of the standard pansharpening toolbox~\cite{vivone2015critical}. The evaluation protocol uses the same MTF convention~\cite{vivone2021benchmark,vivone2021newbenchmark}. Each PAN patch is radiometrically aligned to its LRMS range and broadcast over the spectral channels before the spatial loss, with weights $\eta_{\mathrm{spa}}=\eta_{\mathrm{spe}}=1$ and $\eta_{\mathrm{ori}}=0.1$.

\begin{table}[!t]
\centering
\caption{\textbf{Optimization settings.} Epochs are per target image; each epoch takes $\lceil M^{t}/8\rceil$ steps (later images), 8 steps (seed image), or 1 step (LoRA). LR: learning rate; WD: weight decay.}
\label{tab:training-settings}
\standardtablefont
\setlength{\tabcolsep}{2pt}
\renewcommand{\arraystretch}{1.1}
\begin{tabular*}{\columnwidth}{@{\extracolsep{\fill}}llccr@{}}
\toprule
Study & Schedule & Optimizer & LR & WD \\
\midrule
\multirow{2}{*}{FusionNet/U2Net} & 30 epochs (seed image) & Adam & $10^{-4}$ & $10^{-5}$ \\
& 10 epochs (later images) & Adam & $10^{-4}$ & $10^{-5}$ \\
LoRA & 10 epochs & Adam & $10^{-4}$ & $10^{-5}$ \\
G-ZAP & 500 epochs & Adam & $5\times10^{-4}$ & $10^{-5}$ \\
CrossEarth A2S & 5,000 iterations & AdamW & $10^{-4}$ & $0.05$ \\
\bottomrule
\end{tabular*}
\end{table}

\subsection{LoRA Baseline}
\label{app:lora-settings}

Low-rank branches are attached to the eligible convolutional layers of each backbone (10 for FusionNet, 19 for U2Net); depthwise and transposed convolutions and non-convolutional modules are left unchanged. The rank is $r=\lceil P_{T}/p_{1}\rceil$, where $P_{T}$ is the Feature Tailor parameter count and $p_{1}$ the parameter count of a rank-1 adapter over the injected layers, with scaling factor $\alpha=r$. This gives $r=75$ and $76$ for FusionNet and $r=32$ and $33$ for U2Net on WV3$\rightarrow$WV2 and QB$\rightarrow$GF2, and trainable parameters within 0.2\%--2.4\% of the Tailor size (Table~\ref{tab:lora-control}).

\subsection{Cross-Task Settings}
\label{app:cross-task-settings}

Both tasks apply the residual correction of Eq.~\eqref{eq:tailored-output} at an intermediate interface of a frozen network (Table~\ref{tab:retailor-configurations}) and train only the Tailor, from unlabeled PAN/LRMS observations for G-ZAP and from segmentation labels for CrossEarth.

For G-ZAP, each transfer starts from the source-sensor checkpoint of the original protocol~\cite{yang2026gzap}. The Tailor is adapted from target PAN/LRMS observations and their multiscale degradations under the task's semi-supervised losses, and is inherited along the image sequence. The 500-epoch budget follows the original protocol.

CrossEarth follows the Aerial-to-Satellite (A2S) protocol~\cite{gong2026crossearth,ji2019whu}. Both trained variants start from the same Aerial checkpoint, are trained on the labeled Satellite training set with the same schedule and nearly equal trainable parameters, and are evaluated on the Satellite test set. Head-only updates the original segmentation head, contrasting feature-space residual correction with conventional head adaptation.

\FloatBarrier
\section{Design Choices}
\label{app:design-choices}

\subsection{Backbone Selection}
\label{app:backbone-selection}

We use FusionNet and U2Net as frozen backbones because they are established yet structurally different, a residual detail-injection CNN~\cite{deng2021fusionnet} and a spatial--spectral double-U-Net~\cite{peng2023u2net}, so consistent gains on both reflect the adaptation strategy rather than a single architecture. Both also split into a Feature Extractor and a Channel Mapper, which provides the interface for the Tailor (Table~\ref{tab:retailor-configurations}).

\begin{table}[!t]
\centering
\caption{\textbf{Feature Tailor configurations.} Insertion point, feature width $d$, and trainable parameters for each model.}
\label{tab:retailor-configurations}
\standardtablefont
\setlength{\tabcolsep}{3pt}
\renewcommand{\arraystretch}{1.08}
\begin{tabular*}{\columnwidth}{@{\extracolsep{\fill}}>{\raggedright\arraybackslash}p{0.2\columnwidth}>{\raggedright\arraybackslash}p{0.37\columnwidth}rr@{}}
\toprule
Model/task & Insertion point & Width $d$ & Params \\
\midrule
FusionNet & \texttt{res4} $\rightarrow$ \texttt{conv3} & $32$ & $221{,}472$ \\
U2Net & \texttt{s2block4} $\rightarrow$ \texttt{to\_hrms} & $32$ & $221{,}472$ \\
G-ZAP & INR query $\rightarrow$ decoder & $48$ & $498{,}096$ \\
\mbox{CrossEarth A2S} & pixel decoder $\rightarrow$ Mask2Former head~\cite{cheng2022mask2former} & $256$ & $14{,}158{,}080$ \\
\bottomrule
\end{tabular*}
\end{table}

\subsection{Feature Tailor Insertion Position}
\label{app:ft-placement}

\begin{figure}[pos=h]
    \centering
    \includegraphics[width=\columnwidth]{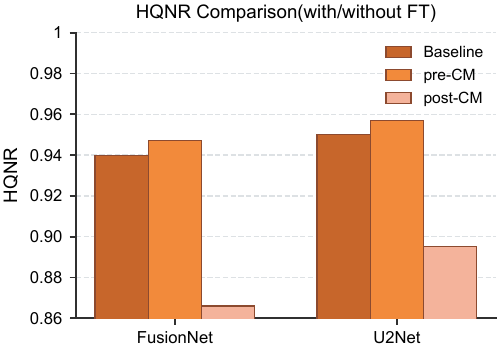}
    \caption{\textbf{HQNR for Tailor placements on WV3.} Baseline: no Tailor; pre-CM/post-CM: before/after the Channel Mapper.}
    \label{fig:ft-placement}
\end{figure}

As defined in Section~\ref{sec:residual-feature-tailoring}, the Tailor sits between $F_{\mathrm{ext}}$ and $F_{\mathrm{map}}$, where the fused feature $\mathbf{Z}$ combines spectral and spatial information from both modalities but is not yet tied to the HRMS space. This agrees with findings that intermediate features transfer better than shallow or output-level ones~\cite{yosinski2014transferable,dou2019domain}. In Fig.~\ref{fig:ft-placement}, inserting the Tailor before the Channel Mapper (pre-CM) gives the highest HQNR on WV3 for both backbones, above both the no-Tailor baseline and insertion after the Channel Mapper (post-CM). Correcting the fused feature is therefore more effective than correcting the output.

\section{Sensitivity to Patch Budget and Patch Size}
\label{app:patch-sensitivity}

\begin{figure}[pos=b]
    \centering
    \setlength{\abovecaptionskip}{4pt}
    \includegraphics[width=0.93\columnwidth]{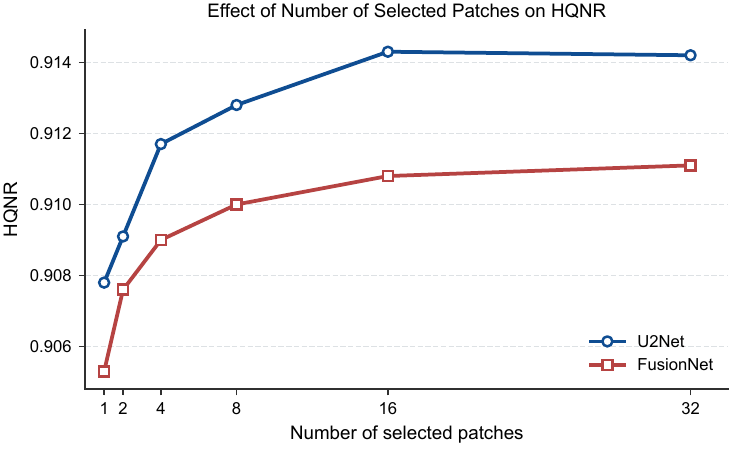}
    \caption{\textbf{HQNR versus number of selected patches (WV3$\rightarrow$WV2).}}
    \label{fig:patch-number-sensitivity}

    \vspace{0.2em}
    \includegraphics[width=0.93\columnwidth]{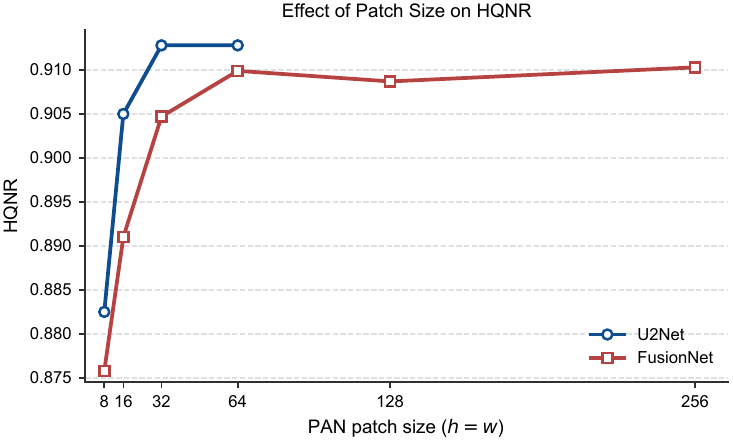}
    \caption{\textbf{HQNR versus PAN patch size (WV3$\rightarrow$WV2).} U2Net is limited to $64\times64$ by GPU memory.}
    \label{fig:patch-size-sensitivity}
\end{figure}

We varied the number of selected patches from 1 to 32 on WV3$\rightarrow$WV2. In Fig.~\ref{fig:patch-number-sensitivity}, HQNR rises markedly up to eight patches for both backbones, while 16 patches add only a modest gain at twice the workload and 32 add almost nothing. We therefore use eight patches as the reference budget.

We also varied the PAN patch size from $8\times8$ up to the largest size that fits in GPU memory. In Fig.~\ref{fig:patch-size-sensitivity}, HQNR rises sharply from $8\times8$ to $32\times32$; at $64\times64$, FusionNet is near its best value and U2Net stays close to its $32\times32$ value. We therefore use $64\times64$, which gives sufficient context at moderate cost.

\section{Complexity and Efficiency of Selection and Tailor Updates}
\label{app:selection-complexity}

\subsection{Asymptotic Complexity}
\label{app:complexity-analysis}

For each later image, selection operates directly on the aligned observations: the two variation cues are computed in parallel for all $N_c$ candidates, followed by at most $M^{t}$ greedy rounds. Its complexity is
\begin{equation*}
    \mathcal{O}\!\left(N_chw+N_cC_{\mathrm{MS}}h_sw_s+M^{t}N_c\right),
\end{equation*}
where $h\times w$ and $h_s\times w_s$ are the PAN and LRMS patch sizes and $C_{\mathrm{MS}}$ is the number of LRMS bands. No backbone forward pass, Tailor update, or pairwise candidate-relation matrix as in~\cite{paul2021datadiet,killamsetty2021glister} is required.

The number of selected patches determines the Tailor-update cost. For a PAN input of size $H\times W$ with $N_c$ patches of size $h\times w$ ($HW=N_chw$) and $E$ adaptation epochs, Table~\ref{tab:adaptation-complexity} compares whole-image adaptation, a fixed budget of $M$ patches, and ReTailor with the image-dependent count $M^{t}$.

\begin{table}[!t]
    \centering
    \caption{\textbf{Complexity of Tailor updates.} (a) Update complexity; (b) compression ratios of ReTailor (attention assumes global self-attention).}
    \label{tab:adaptation-complexity}
    \standardtablefont
    \setlength{\tabcolsep}{4pt}
    \renewcommand{\arraystretch}{1.12}
    \begin{tabular*}{\columnwidth}{@{\extracolsep{\fill}}lccc@{}}
        \toprule
        (a) & Whole image & Fixed ($M$) & ReTailor ($M^{t}$) \\
        \midrule
        CNN & $\mathcal{O}(EHW)$ & $\mathcal{O}(EMhw)$ & $\mathcal{O}(EM^{t}hw)$ \\
        Attention & $\mathcal{O}\!\left(E(HW)^2\right)$ & $\mathcal{O}\!\left(EM(hw)^2\right)$ & $\mathcal{O}\!\left(EM^{t}(hw)^2\right)$ \\
        \midrule
        (b) & \multicolumn{1}{c}{} & Whole/ReTailor & Fixed/ReTailor \\
        \midrule
        CNN & & $\dfrac{N_c}{M^{t}}$ & $\dfrac{M}{M^{t}}$ \\
        Attention & & $\dfrac{N_c^2}{M^{t}}$ & $\dfrac{M}{M^{t}}$ \\
        \bottomrule
    \end{tabular*}
\end{table}

Selecting on average 6.05 and 6.47 of 64 candidates reduces the Tailor-update workload by 90.5\% and 89.9\% relative to whole-image adaptation on WV2 and GF2, and the measured adaptation time by 9.9\%--17.1\% relative to fixed-8 adaptation, including selection overhead.

\subsection{Efficiency of the Tensorized Greedy Scan}
\label{app:tensorized-greedy-scan}

\begin{table}[!t]
\centering
\caption{\textbf{Selector latency on QB$\rightarrow$GF2.} Time of a single selection, mean $\pm$ std over images.}
\label{tab:tensorized-greedy-runtime}
\standardtablefont
\setlength{\tabcolsep}{3pt}
\renewcommand{\arraystretch}{1.08}
\begin{tabular*}{\columnwidth}{@{\extracolsep{\fill}}lcc@{}}
\toprule
Candidate evaluation & Latency (ms) $\downarrow$ & Speedup $\uparrow$ \\
\midrule
Sequential & $5.28\pm0.84$ & $1.00\times$ \\
Parallel & $\mathbf{1.99\pm0.28}$ & $\mathbf{2.65\times}$ \\
\bottomrule
\end{tabular*}
\end{table}

We compared the tensorized selector with a functionally equivalent Python-loop variant that uses identical scores, regions, and stopping rule but evaluates candidates sequentially within each round. On the post-seed QB$\rightarrow$GF2 images (64 candidates each, RTX 4090), each selector was run 20 times after warm-up per image. Table~\ref{tab:tensorized-greedy-runtime} shows that tensorization reduces the latency of a single selection from $5.28\pm0.84$ ms to $1.99\pm0.28$ ms (62.3\% lower, $2.65\times$ faster) while producing identical selections.

\FloatBarrier
\renewcommand{\bibfont}{\fontsize{9pt}{10.5pt}\selectfont}
\setlength{\bibsep}{1pt}
\bibliographystyle{elsarticle-num}
\bibliography{reference}
\end{document}